\documentclass[letterpaper]{article} 
\usepackage{aaai25}  
\usepackage{times}  
\usepackage{helvet}  
\usepackage{courier}  
\usepackage[hyphens]{url}  
\usepackage{graphicx} 
\usepackage{natbib}  
\usepackage{caption} 
\usepackage{algorithm}
\usepackage{algorithmic}

\usepackage{newfloat}
\usepackage{listings}
\DeclareCaptionStyle{ruled}{labelfont=normalfont,labelsep=colon,strut=off} 
\floatstyle{ruled}
\newfloat{listing}{tb}{lst}{}
\floatname{listing}{Listing}

\title{\textsc{SMoSE}: Sparse Mixture of Shallow Experts for\\Interpretable Reinforcement Learning in Continuous Control Tasks}

\author{
 Mátyás Vincze\textsuperscript{\rm 1,2},
 Laura Ferrarotti\textsuperscript{\rm 2},
 Leonardo Lucio Custode\textsuperscript{\rm 1}\\
 Bruno Lepri\textsuperscript{\rm 2},
 Giovanni Iacca\textsuperscript{\rm 1}
}

\affiliations{
 \textsuperscript{\rm 1} University of Trento, Italy\\
 \textsuperscript{\rm 2} Fondazione Bruno Kessler, Italy\\
 {\rm \{mvincze, ferrarotti, lepri\}@fbk.eu, \{leonardo.custode, giovanni.iacca\}@unitn.it}
}

\usepackage{comment}
\usepackage{xspace}
\usepackage{xcolor}
\usepackage{amsmath}
\usepackage{amsfonts}
\usepackage{amsthm}
\usepackage{svg}

\usepackage{graphicx} 
\usepackage{subcaption}

\usepackage{booktabs} 
\usepackage{multirow}

\newtheorem*{remark}{Remark}

\newcommand{\algname}{\textsc{SMoSE}\xspace}
\newcommand{\best}[1]{{\color{magenta}{#1}}}

\begin{document}

\maketitle

\begin{abstract}
Continuous control tasks often involve high-dimensional, dynamic, and non-linear environments. State-of-the-art performance in these tasks is achieved through complex closed-box policies that are effective, but suffer from an inherent opacity. Interpretable policies, while generally underperforming compared to their closed-box counterparts, advantageously facilitate transparent decision-making within automated systems. Hence, their usage is often essential for diagnosing and mitigating errors, supporting ethical and legal accountability, and fostering trust among stakeholders. In this paper, we propose \algname, a novel method to train sparsely activated interpretable controllers, based on a \mbox{\emph{top}-$1$} Mixture-of-Experts architecture. \algname combines a set of interpretable decision-makers, trained to be experts in different basic skills, and an interpretable router that assigns tasks among the experts. The training is carried out via state-of-the-art Reinforcement Learning algorithms, exploiting load-balancing techniques to ensure fair expert usage. We then distill decision trees from the weights of the router, significantly improving the ease of interpretation. We evaluate \algname on six benchmark environments from MuJoCo: our method outperforms recent interpretable baselines and narrows the gap with non-interpretable state-of-the-art algorithms.
\end{abstract}

\section{Introduction} 

Over the last decade, Artificial Intelligence (AI) has achieved dramatic success. However, the increasing adoption of AI systems in real-world use cases has been raising concerns related to the trustworthiness of these systems \cite{wexler2017computer, mcgough2018bad, varshney2017safety, rudin_age_2019, huang2020survey, smyth2021understanding,he2021liable}. One of the most promising paths toward trustworthy AI involves developing methods to enhance our understanding of AI decision-making processes. In this context, eXplainable AI (XAI) introduces AI systems enabling the generation of post-hoc explanations for their behavior \cite{dwivedi2023explainable}. Although such methods do provide insights into the decision-making processes, their post-hoc analysis explains just an approximation of the original closed-box behavior. This approximation cannot be trusted in high-stakes scenarios, which are frequent for instance in robot control, autonomous driving, emergency response, and public decision-making. Indeed, given that interpretable policies distilled for explanation are typically less complex than closed-box ones, they might require adopting entirely different strategies, disrupting the link between the original behavior and its interpretation \cite{rudin_stop_2019}. To overcome this issue, research efforts have been channeled towards Interpretable AI (IAI) \cite{rudin_stop_2019, akrour2021continuous}. Interpretable AI focuses on systems whose decision-making process is inherently understandable to humans, without the need for distilled explanations \cite{arrieta2020explainable}. These models prioritize clarity and simplicity, enabling users to directly comprehend how inputs are transformed into outputs, thus facilitating transparency, trust, and validation ease \cite{rudin_stop_2019}.

Due to its sequential nature, Reinforcement Learning (RL) benefits more from directly interpretable policies than post-hoc explanations \cite{rudin_interpretable_2021}. Discrepancies between a policy and its interpretable approximation can grow over time due to state distribution drift, making local explanations less meaningful \cite{akrour2021continuous}. Moreover, post hoc XAI methods, when used in RL to tackle goal misalignment problems, might mislead observers into believing that closed-box agents select the correct actions for the appropriate reasons, even though their decision-making process may actually be misaligned \cite{chan2022comparativestudyfaithfulnessmetrics, delfosse2024interpretableconceptbottlenecksalign}. 
This paper addresses the challenge of Interpretable Reinforcement Learning (IRL) in continuous action spaces. 
In particular, the main advances presented in this work can be summarized as follows: 

\begin{itemize}
    \item We propose \algname, a novel, high-performing method, that exploits a sparse, interpretable Mixture-of-Experts (MoE) architecture. Our method simultaneously learns a set of simple yet effective continuous sub-policies, and an interpretable router. The sub-policies are sequentially selected by the router, one per each decision step, to control the system, based on the current state. 
    \item We showcase the capabilities of \algname through results obtained on six well-known continuous control benchmarks from the MuJoCo environment \cite{MuJoCo}. For all the considered environments we analyze performance both in training and in evaluation. 
    \item We include a full interpretation for all the trained policies, demonstrating the effectiveness of \algname in providing extensive insight on the learned controllers. By combining router and the experts interpretation, we analyze the high-level and low-level alignment of the policy.
    \item We compare \algname with interpretable and non-interpretable state-of-the-art RL models, considering both larger model size competitors, and models with comparable size (in terms of the number of active and overall parameters). Results highlight that the proposed architecture, while retaining interpretability, improves the performance w.r.t. other interpretable methods, tightening the gap with non-interpretable approaches.
\end{itemize}

The paper is structured as follows: the next two sections summarize the relevant background and related contributions, while the Methods section details \algname, our proposed approach. Then, the Results section presents the experimental setup and the performance achieved by our method, including the interpretation of the learned policies. Finally, we draw the conclusions and discuss the future directions of this work.

\begin{figure*}[ht!]
    \centering
    \includegraphics[width=0.85\textwidth]{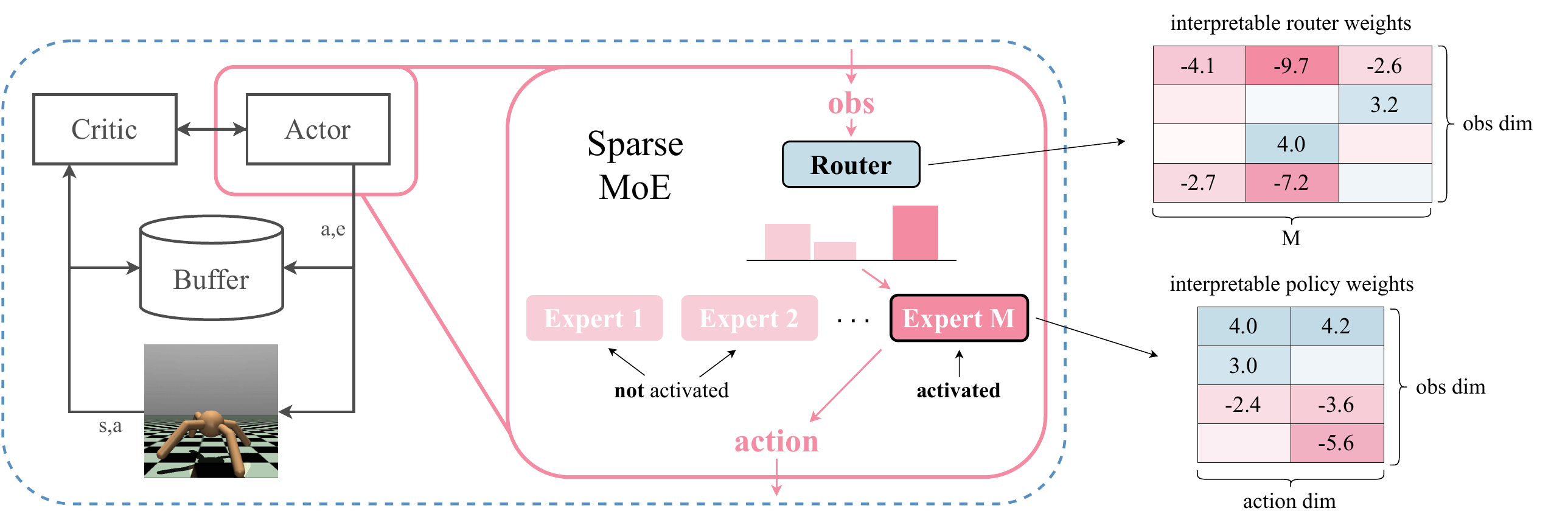}
    \caption{\textbf{\algname}. Schematic summary of the proposed architecture.}
    \label{fig:smose_main_plot}
\end{figure*}

\section{Background}

As recently surveyed in \cite{glanois2024survey}, several works study IRL models. 
Some approaches exploit neural logic-based policies to achieve interpretability \cite{jiang2019neurallogicreinforcementlearning, kimura2021neurosymbolicreinforcementlearningfirstorder, delfosse2023interpretableexplainablelogicalpolicies, sha2024expilexplanatorypredicateinvention}.
Older approaches \cite{mccallum1996reinforcement, pyeatt_computer_nodate} use existing methods for decision tree (DT) induction, adapting them to the RL domain. In \cite{roth2019conservative}, the authors introduce a heuristic to keep the size of the trees smaller while still achieving good performance. However, these algorithms suffer from the curse of dimensionality, i.e., they do not scale well with the dimensionality of the state space. More recent approaches address this issue. 
In \cite{silva_optimization_2020}, the authors employ soft DTs \cite{irsoy2012soft} as policies for RL agents. This simplifies training and allows the use of widely known deep RL algorithms. However, soft trees are difficult to interpret, and discretizing them into ``hard'' DTs, the policies obtained can suffer from a significant loss in performance. 

Other approaches make use of evolutionary principles to optimize DTs. \cite{dhebar_interpretable-ai_2020} propose a method for learning a non-linear DT from a neural network trained on the target environment. This allows for choosing the desired properties of the resulting DT (e.g., the depth). On the other hand, this methodology hinders online learning (and thus adaptation to novel scenarios). In \cite{custode2023evolutionary}, the authors combine evolutionary techniques with $Q$-Learning to produce DTs that can learn online, while still being interpretable. The DTs produced by this approach achieve performance comparable to non-interpretable state-of-the-art algorithms in a number of simple benchmarks. However, this approach has an extremely high computational cost, requiring a large number of interactions with the environment. Finally, in \cite{custode_sirl}, the authors leverage principles from social learning to significantly improve both the computational complexity and the performance of evolutionary methods.

\section{Related work}

IRL methods tailored to environments with continuous action space are heavily needed in a wide variety of real-world scenarios, e.g., robot manipulation and control, as showcased by many benchmarking examples \cite{MuJoCo}. So far, however, only a few works have investigated the use of IRL for continuous control. A branch of research is dedicated to the learning of interpretable programs as RL policies \cite{verma2019programmaticallyinterpretablereinforcementlearning, verma2021imitationprojectedprogrammaticreinforcementlearning, liu2023hierarchicalprogrammaticreinforcementlearning, kohler2024interpretableeditableprogrammatictree}. In \cite{custode_co_2021} the authors propose a cooperative co-evolutionary approach in order to independently evolve a population of binary DTs (generated via Grammatical Evolution) and a population of sets of actions, both optimized w.r.t. the fitness associated with the combined use of the two. \cite{videau2022multi} explore methods for constructing symbolic RL controllers, utilizing parse trees and Linear Genetic Programming (LGP) to represent the programs as a vector of integers. Additionally, a multi-armed bandit strategy distributes the computational budget across generations. LGP is also used in \cite{nadizar2024naturally}, along with Cartesian Genetic Programming (CGP), where programs are instead represented as directed acyclic graphs. In \cite{paleja2023interpretable}, an IRL algorithm for continuous control is introduced, exploiting fuzzy DTs combined with nodes and leaves with differentiable crispification, that can hence be directly learned via gradient descent. As previously mentioned, our method takes advantage of an interpretable MoE architecture for the control policy. MoEs can be found in RL literature, employed to tackle different problems, such as parameter scaling in deep RL \cite{obando2024mixtures}, handling of multiple optimal behaviors \cite{ren2021probabilistic}, and multi-task learning \cite{cheng2023multi, willi2024mixtureexpertsmixturerl}, to name a few. In the realm of interpretability, a kernel-based method employing MoE is proposed in \cite{akrour2021continuous}. In this work, the selection of a set of centroids from trajectory data is optimized. Each state is associated with an expert policy modeled as a Gaussian distribution around a linear policy, while retaining an internal complex function approximator. According to this approach, a learned combination of experts handles the control task. This is obtained by considering fuzzy memberships to clusters and employing a learned set of cluster weights. In our method, instead, we can tune the number of experts employed at every timestep for the decision, and, for instance, force the policy to exploit only one expert, for maximum interpretability (as explained in more detail in the Methods section). 
Moreover, while the policies in \cite{akrour2021continuous} are updated via approximate policy iteration, the centroids, which must be elements of the replay buffer, require separate iterations of discrete optimization. \algname, instead, learns a router that distributes the control tasks among experts. This compact representation, while being fully interpretable, can be learned via backpropagation, simultaneously to the experts.

\section{Method}
We seek to solve RL problems with continuous actions structured as Markov Decision Processes (MDPs), i.e., tuples $(\mathcal{S}, \mathcal{A}, \mathcal{P}, \mathcal{R}, \gamma, \mathcal{S}_0)$, where $\mathcal{S}$ is the set of the states in the problem, $\mathcal{A} \in \mathbb{R}^{n_a}$ is the set of (continuous) actions, $\mathcal{P}(s, a, s'): \mathcal{S} \times \mathcal{A} \times \mathcal{S} \rightarrow [\, 0, \, 1 \, ]$ associates a probability to each transition from $(s, a)$ to each state $s'$;  $\mathcal{R}(s, a, s'): \mathcal{S} \times \mathcal{A} \times \mathcal{S} \rightarrow \mathbb{R}^{+}$ assigns a reward to each triplet $(s, a, s')$; $\gamma$ is a discount factor, used for denoting the importance of future rewards w.r.t. the current one, and $\mathcal{S}_0$ is the set of the initial states. Solving such a problem requires the design of a learning strategy to fit a policy function $\pi: \mathcal{S} \times \mathcal{A} \rightarrow [\, 0, \, 1 \, ]$, optimizing: 
\begin{equation*}
\mathbb{E}_{a_t \sim \pi_t}\Big[  \sum_{t=0}^{\infty} \gamma^{t} \mathcal{R}_t \, \Big\vert \, s_0 \sim \mathcal{S}_0, \, s_{t+1} \sim \mathcal{P}_t \, \Big]\footnote{$\pi_t \doteq \pi(s_t, \cdot \, )$, $\mathcal{R}_t \doteq \mathcal{R}(s_t, a_t, s_{t+1})$, $\mathcal{P}_t \doteq \mathcal{P}(s_t, a_t, \cdot \, )$ }
\end{equation*}
through interaction with the environment. Good results tackling this problem in non-trivial environments are often achieved by complex, not directly interpretable policies.

The main idea behind our method is to decompose a complex behavior by identifying a set of basic skills (or decision strategies) that are easier to interpret when combined with a policy capable of selecting which skill to employ in each state. Decomposing the decision process with this  \emph{divide-and-conquer} approach can provide insight into the interpretation of the learned controller's behavior. Thus, we structure the control policy as a composition of:
\begin{itemize}
    \item A set $\lbrace \pi_m(\cdot \, \vert \, \theta_m):\mathcal{S} \rightarrow \mathcal{A} \,\, \vert \,\, m = 1, \dots, M \rbrace$ of $M$ parameterized continuous policy functions, representing $M$ decision-makers, each trained to be expert in an individual useful basic skill. To ensure overall interpretability, we consider continuous policies $\pi_m$ that are shallow and directly interpretable, such as linear ones. 
    \item A planner, capable of assigning to each state $s\in\mathcal{S}$ an expert policy among the available set, in order to attain optimal behavior. This module is a parameterized router $g(\, s \, \vert \, \Theta \, ):\mathcal{S} \rightarrow [\, 0 \, , \, 1 \, ]^M$ producing a one-hot encoded $M$-dimensional vector as output. 
\end{itemize}

\algname, summarized in Figure \ref{fig:smose_main_plot}, combines the experts and the router according to a MoE architecture of the form:
\begin{equation}
  \pi(\, s \, )= \sum_{m=1}^{M} \, \left[ \, g(\, s \, \vert \, \Theta \, ) \, \right]_{m} \,\pi_m(\, s \, \vert \, \theta_m \, ), 
  \label{eq:moe}
\end{equation}
where $\left[ \, g(\, s \, \vert \, \Theta \, ) \, \right]_{m}$ is the $m$-th component of vector:
\begin{equation}
g(\, s \, \vert \, \Theta \, ) = \text{TOP}_1(\, \text{softmax}(\, \hat{g}(\, s \, \vert \, \Theta\,)\,).
\label{eq:router}
\end{equation}
In the equation above, we indicate as $\hat{g}(\, s \, \vert \, \Theta \,) \in \mathbb{R}^M$ the inner parameterization of the router that produces as output an $M$-dimensional vector. The softmax function then transforms $\hat{g}(\, s \, \vert \, \Theta \,)$ in a preference vector, where the $m$-th element measures the preference in choosing the $m$-th expert as actor in state $s$. The function $\text{TOP}_1$ assigns value $1$ to the vector component with maximum preference, and $0$ to all the others, allowing for an extremely sparse MoE structure in which only one expert is activated in each state. 
\begin{remark}
The proposed method can also accommodate the use of $\text{TOP}_k$ with $k>1$, replacing $\text{TOP}_1$ in Eq. \eqref{eq:router} and thereby enabling the selection of a combination of $k$ experts at each timestep to make control decisions. While the resulting policy would remain interpretable, in this work we intentionally set $k=1$ to evaluate the performance of a truly sparse MoE architecture and to maximize interpretability within this framework.
\end{remark}
This architecture takes inspiration from \citep{shazeer2017outrageouslylargeneuralnetworks, MoEV}, where sparse MoE neural layers are stacked to obtain both computational efficiency in inference and parameters specialized on subsets of states. Here, such architecture is tuned to retain interpretability in continuous control, by removing the non-linearities and activation functions, employing a single-layer shallow structure, and shaping the inner router $\hat{g}$ and the experts $\pi_m$ as linear functions, that is:
\begin{equation*}
 \pi_m(\, s \, \vert \, \theta_m \, ) = \theta_m \cdot s \quad \textrm{and} \quad \hat{g}(\, s \, \vert \, \Theta \, ) =  \Theta \cdot s
\end{equation*}
with $\theta_m \in \mathbb{R}^{n_a \times n_s}$ and $\Theta \in \mathbb{R}^{M \times n_s}$.

The control policy in Eq. \eqref{eq:moe} is trained via RL. The parameters characterizing $\pi_m$ and $g$ are hence simultaneously learned. It is important to notice that composing the $\text{TOP}_1$ and softmax functions in the order expressed in Eq. \eqref{eq:router} permits a larger propagation of the gradients among the router weights, if compared with the opposite ordering. At every update, indeed, each of the $M$ components of $\text{softmax}(\, \hat{g}(\, s \, \vert \, \Theta\,) \, )$ depends on all the weights $\Theta$, due to the action of softmax. Hence, all the components in $\Theta$ will be updated, not only the ones associated with the expert selected by $\text{TOP}_1$ \citep{MoEV}. We ensure to explore the action space of each expert $\pi_m$ by injecting stochasticity in the choices in training, i.e., $$\pi_m(\, s \, \vert \, \theta_m, \sigma_m \, ) = \mathcal{N}(\, \theta_m \cdot s, \,\,\,\, \sigma_m^2 \, ).$$

Our method can be seamlessly integrated with any RL algorithm for the learning of continuous controllers. In this work, we rely for exploration on Soft Actor-Critic (SAC) \cite{SAC}, a state-of-the-art algorithm that balances the objective function with a term promoting higher entropy policies. We structure the actor module according to the interpretable architecture in Eq. \eqref{eq:moe}, while we maintain the classic neural critic introduced in \cite{SAC}. 

In order to ensure balanced workloads among the $M$ experts, we augment the SAC actor objective function with additional penalties introduced in \citep{MoEV}, weighted by a tunable parameter $\lambda$. In particular, per each mini-batch $\rm{S} = \lbrace s_k \rbrace \subseteq \mathcal{S} $ of states, we consider its \emph{importance} for $m = 1, \dots , M$:
\begin{align*}
\text{Imp}_m({\rm S}) = \sum_{s_{\rm k} \in {\rm S}} \text{softmax}(\pi_m(\, s_{\rm k}\, \vert \, \theta_m , \sigma_m \, )) 
\end{align*}
and we compute the \emph{importance loss}:
\begin{align}
f_{\text{imp}}({\rm S}) = \frac{1}{2} \Big(\frac{\text{std}(\text{Imp}({\rm S}))}{\text{mean}(\text{Imp}({\rm S}))}\Big)^2.
\label{eq:imp-loss}
\end{align}
Then, we consider the \emph{load} of $\rm{S}$ for $m = 1, \dots , M$:
\begin{equation*}
\text{Load}_m({\rm S}) = \sum_{s_{\rm k} \in {\rm S}} \mathbb{P}( \epsilon_{\text{new}} \geq \tau(s_{\rm k}) - \pi_m(\, s_{\rm k} \, \vert \, \theta_m , \sigma_m \, ))
\end{equation*}
with $\tau(s_{\rm k}) = \text{max}_m \Big(\pi_m(\, s_{\rm k} \, \vert \, \theta_m , \sigma_m \, )  \Big)$, and compute the \emph{load-balancing loss}:
\begin{equation}
f_{\text{load}}({\rm S}) = \frac{1}{2} \Big(\frac{\text{std}(\text{Load}({\rm S}))}{\text{mean}(\text{Load}({\rm S}))}\Big)^2.
\label{eq:load-loss}
\end{equation}
When computing the load-balancing loss on each mini-batch $\rm{S}$, noise to the inner router, i.e., $ \hat{g}(\, s \, \vert \, \Theta \, ) =  \Theta \cdot s + \epsilon $, with $\epsilon \sim \mathcal{N}(\, 0, \, 1/M^2 \, )$, in order to ensure proper exploration of experts employment. Finally, once the policy training is complete, we produce a useful support for interpretation by distilling a multiclass classifier using Decision Trees (DTs), which are trained on a router-labeled replay buffer. By decomposing the original router into $M$ binary classifiers of limited depth, we create an easily readable representation of the interpretable router's decision-making process, where each DT determines whether a specific expert should control the task for a given state. After balancing the data, we train the DTs in a supervised manner, using the CART algorithm \cite{timofeev2004classification}. More details on this can be found in the Appendix.

\section{Results}

In this section, we present the evaluation of \algname on six widely-known continuous-control environments from MuJoCo \cite{MuJoCo}, namely Walker2d-v4, Hopper-v4, Ant-v4, HalfCheetah-v4, Reacher-v4, and Swimmer-v4 (described in Appendix), which are commonly used as benchmarks in the RL field.
The following subsections detail the experimental setup, performance evaluation, and comparative analysis with interpretable and non-interpretable baselines, highlighting the effectiveness of \algname in delivering both interpretability and high performance. 
As it is common in RL literature, performance is measured considering the attained episodic rewards (ER), i.e., the accumulated rewards achieved at each timestep of an episode, in order to compare well with the state of the art.

\subsection{Training performance}

\begin{figure}[t]
\centering
\includegraphics[width=1\columnwidth]{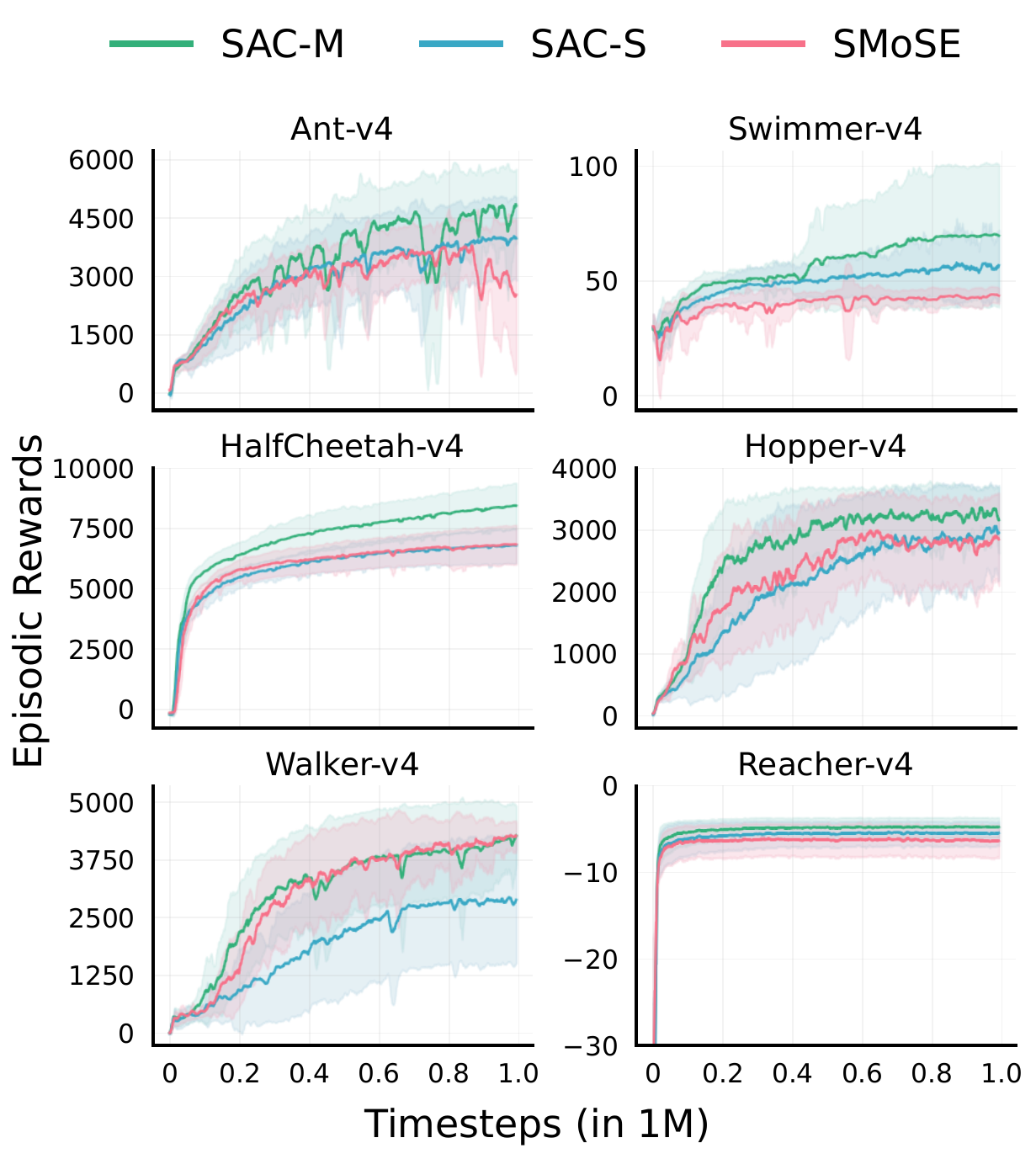}
\caption{\textbf{Performance in training.} \algname compares to non-interpretable models of the same size, considering the number of overall (SAC-M) and active (SAC-S) parameters.}
\label{fig:performance-training}
\end{figure}

In this section, we include the results in terms of performance achieved in training by \algname, tuned with $M=8$ experts, and weighing the load-balancing losses in Eq.s \eqref{eq:imp-loss}-\eqref{eq:load-loss} with $\lambda = 0.1$. The value for the $M$ parameter has been empirically tuned to strike a balance between agents' performance and easiness of interpretation, as shown in the ablation study included in Appendix. 
The weight $\lambda$ was tuned between $0.01$ and $1.0$ on a three-dimensional grid, to ensure fair load and reduce expert collapse. Details on the computational setup are available in the Appendix. To achieve statistical reliability in our results, we perform $N_{\text{train}} = 10$ independent training runs, seeded from 0 to 9. Every training run consists of a sequence of episodes with a maximum horizon of $H = 1000$ interactions with the environment, which is achieved if the controller does not incur catastrophic failure.

\begin{table}[ht!]
 \centering
  \caption{\textbf{Performance and model size comparison}. \algname outperforms interpretable methods.}
  \label{tab:eval_interpretable}
  \resizebox{\columnwidth}{!}{
  \begin{tabular}{lllc}
  \toprule
  \textbf{Environment} & \textbf{Algorithm} & \multicolumn{1}{c}{\textbf{avg. ER}} & $\text{\textbf{N}}_{\text{\textbf{act}}} (\text{\textbf{N}}_{\text{\textbf{tot}}})$ \\
  \midrule  
  \multirow{4}{*}{Walker2d-v4} & CGP & $\phantom{-}1090.00^* \pm \phantom{0}59.50^*$  & ---  \\
                               & LGP & $\phantom{-}1080.00^* \pm \phantom{0}14.00^*$ & ---  \\
                               & Metric-40 & $\phantom{-0}775.00^* \pm 115.50^*$ & $1006$ \\
                               & \algname (ours)  & \best{$\phantom{-}4224.29\phantom{*} \pm \phantom{0}25.96$} & $360$ ($1872$)  \\
  \midrule
  \multirow{4}{*}{Hopper-v4}  & CGP & $\phantom{-}1150.00^* \pm \phantom{0}92.50^*$  & ---  \\
                              & LGP & $\phantom{-}1120.00^* \pm \phantom{0}87.50^*$ & ---  \\
                              & Metric-40 & $\phantom{-}2005.00^* \pm 295.00^*$  & $643$ \\
                              & \algname (ours) & \best{$\phantom{-}2816.08\phantom{*} \pm 445.57$} & $168$ ($672$)  \\
  \midrule
  \multirow{4}{*}{Ant-v4}     & CGP & $\phantom{-}1130.00^* \pm 222.50^*$ & ---  \\
                              & LGP & $\phantom{-}1210.00^* \pm 390.00^*$ & ---  \\
                              & Metric-40 & $\phantom{-}2210.50^* \pm 175.50^*$ & $1488$  \\
                              & \algname (ours) & \best{$\phantom{-}3245.43\phantom{*} \pm 380.93$}  & $672$ ($3808$)  \\
  \midrule
  \multirow{4}{*}{HalfCheetah-v4} & CGP & $\phantom{-}6375.00^* \pm 496.50^*$  & --- \\
                                  & LGP & $\phantom{-}6388.50^* \pm 296.50^*$ & ---  \\
                                  & Metric-40 & $\phantom{-}2210.50^* \pm 175.50^*$  & $1006$ \\
                                  & \algname (ours) & \best{$\phantom{-}7310.17\phantom{*} \pm 131.57$}  & $360$ ($1872$)  \\
  \midrule
  \multirow{3}{*}{Reacher-v4} & CGP & $\phantom{0}-68.50^* \pm \phantom{0}43.75^*$ & --- \\
                              & LGP & $\phantom{0}-58.50^* \pm \phantom{0}11.10^*$  & ---  \\
                              & \algname (ours) & \best{$\phantom{00}-5.49\phantom{*} \pm \phantom{00}2.32$} & $360$ ($1872$) \\
  \midrule
  \multirow{3}{*}{Swimmer-v4} & CGP & \best{$\phantom{-0}280.00^* \pm \phantom{00}7.50^*$}  & --- \\
                              & LGP & $\phantom{-0}278.50^* \pm \phantom{0}14.00^*$ & ---  \\
                              & \algname (ours) & $\phantom{-00}45.40\phantom{*} \pm \phantom{00}1.62$  & $108$ ($360$)  \\
  \midrule 
  \midrule
  \multicolumn{4}{l}{`*' = visually derived from the plots reported in the original papers} \\
  \multicolumn{4}{l}{`---' = number of employed parameters not specified in literature} \\
  \multicolumn{4}{l}{\best{magenta} = best score per environment}  \\
  \bottomrule
  \end{tabular}%
 }
\end{table}

We train for a total of one million environmental interactions (i.e., timesteps), to be comparable with the closed-box methods literature. The ER achieved by \algname in training over the six mentioned environments is represented in Figure \ref{fig:performance-training}, where we plot the mean and standard deviation of episodic returns over the environment interactions. We assign the ER to all the timesteps of the same episode, and then we plot the ER at each timestep, to make episodes with different lengths comparable. 
In the same figure, the ER achieved by closed-box policies is included as a reference. For a fair comparison, we consider neural policies trained with SAC (the same RL algorithm we use to learn our policy), tuned with the same set of hyperparameters, included in the Appendix. Moreover, we consider non-linear structures that exploit a comparable amount of parameters with our interpretable policy. In particular, our model has a total number $N_{\text{tot}}$ of parameters, considering both the router and the $M$ experts; however, for each decision, only $N_{\text{act}}$ active parameters are used (i.e., those associated with the router and the selected expert), which alleviates the computational cost of the decision-making process. The figure includes the results in training associated with a first neural architecture, indicated as SAC-M, that exploits $N_{\text{tot}}$ parameters for each decision. Additionally, the figure shows as well the results of a second neural architecture, indicated as SAC-S, that exploits $N_{\text{act}}$ parameters for each decision, and hence fully comparable in inference to our policy. Both $N_{\text{tot}}$ and $N_{\text{act}}$ for our method are detailed 
in Tables \ref{tab:eval_interpretable}-\ref{tab:eval_non_interpretable}. From Figure \ref{fig:performance-training}, we can see that \algname's performance is close to the one of SAC-S on three over six environments (HalfCheetah-v4, Hopper-v4, and Reacher-v4), and sensibly better than it on one over six environments (Walker2d-v4), while it is close to the result achieved by SAC-M on two over six environments (Walker2d-v4 and Reacher-v4). These evaluations are affected by the performance on Ant-v4, where we note that the initial behavior, similar again to the one of SAC-S, is worsened by a small number of final under-average training realizations ($3$ over $10$), while higher performance is achieved in the majority of the cases.

\begin{table}[t!]
 \centering
  \caption{\textbf{Performance and model size comparison}. \algname narrows the gap with non-interpretable methods.}
 \label{tab:eval_non_interpretable}
 \resizebox{\columnwidth}{!}{
  \begin{tabular}{lllc}
  \toprule
  \textbf{Environment} & \textbf{Algorithm} & \multicolumn{1}{c}{\textbf{avg. ER}} & $\text{\textbf{N}}_{\text{\textbf{act}}} (\text{\textbf{N}}_{\text{\textbf{tot}}})$  \\
  \midrule
  \multirow{5}{*}{Walker2d-v4} & SAC-L & \best{$\phantom{-0}4358.06 \pm \phantom{0}582.94$}  & $73484$ \\
  & SAC-M  & $\phantom{-0}4020.51 \pm \phantom{0}192.75$ & $1842$ \\
  & SAC-S  & $\phantom{-0}2967.14 \pm \phantom{00}77.18$ & $372$ \\
  & PPO & $\phantom{-0}3362.16 \pm \phantom{0}793.40$ & $5708$ \\
  & \algname (ours) & $\phantom{-0}4224.29 \pm \phantom{00}25.96$ & $360$ ($1872$) \\
  \midrule
  \multirow{5}{*}{Hopper-v4}  & SAC-L & $\phantom{-0}2636.49 \pm \phantom{0}424.21$  & $70406$ \\
  & SAC-M  & \best{$\phantom{-0}3224.25 \pm \phantom{0}177.90$} & $672$ \\
  & SAC-S  & $\phantom{-0}3076.09 \pm \phantom{0}178.37$ & $186$ \\
  & PPO & $\phantom{-0}2311.90 \pm \phantom{0}654.90$ & $5126$ \\
  & \algname (ours) & $\phantom{-0}2816.08 \pm \phantom{0}445.57$ & $168$ ($672$) \\
  \midrule
  \multirow{5}{*}{Ant-v4}  & SAC-L & \best{$\phantom{-0}5255.46 \pm 1070.65$} & $77072$ \\
  & SAC-M  & $\phantom{-0}4894.18 \pm \phantom{0}599.64$  & $3800$ \\
  & SAC-S  & $\phantom{-0}4162.97 \pm \phantom{0}298.12$  & $720$ \\
  & PPO & $\phantom{-0}2327.12 \pm \phantom{0}871.63$  & $6480$ \\
  & \algname (ours) & $\phantom{-0}3245.43 \pm \phantom{0}380.93$  & $672$ ($3808$) \\
  \midrule
  \multirow{5}{*}{HalfCheetah-v4} & SAC-L & \best{$\phantom{-}11809.87 \pm \phantom{0}256.10$} & $73484$ \\
  & SAC-M  & $\phantom{-0}8992.22 \pm \phantom{00}80.58$ & $1842$ \\
  & SAC-S  & $\phantom{-0}7214.30 \pm \phantom{00}87.29$ & $372$ \\
  & PPO & $\phantom{-0}2308.29 \pm 1526.87$ & $5708$ \\
  & \algname (ours) & $\phantom{-0}7310.17 \pm \phantom{0}131.57$  & $360$ ($1872$) \\
  \midrule
  \multirow{5}{*}{Reacher-v4}  & SAC-L & \best{$\phantom{000}-3.75 \pm \phantom{000}1.50$} & $69892$ \\
  & SAC-M & $\phantom{000}-4.02 \pm \phantom{000}1.61$  & $484$ \\
  & SAC-S  & $\phantom{000}-4.82 \pm \phantom{000}1.81$  & $148$ \\
  & PPO & $\phantom{000}-6.57 \pm \phantom{000}2.37$  & $4930$ \\
  & \algname (ours) & $\phantom{000}-5.49 \pm \phantom{000}2.32$  & $144$ ($480$) \\
  \midrule
  \multirow{5}{*}{Swimmer-v4}  & SAC-L & $\phantom{-000}68.59 \pm \phantom{000}2.87$  & $69124$ \\
  & SAC-M  & $\phantom{-000}71.94 \pm \phantom{000}1.89$  & $355$ \\
  & SAC-S  & $\phantom{-000}59.42 \pm \phantom{000}2.89$  & $108$ \\
  & PPO & \best{$\phantom{-000}93.26 \pm \phantom{00}19.90$}  & $4868$ \\
  & \algname (ours) & $\phantom{-0000}45.4 \pm \phantom{000}1.62$ & $108$ ($360$) \\
  \midrule
  \midrule
  \multicolumn{4}{l}{\best{magenta} = best score per environment}  \\
  \bottomrule
  \end{tabular}%
 }
\end{table}

\subsection{Policy evaluation}
After training, we evaluate the performance of the learned policies in Eq. \eqref{eq:moe} once deployed, and compare it with interpretable and not-interpretable methods on the six environments. We test deterministic linear experts obtained discarding the learned standard deviations $\sigma_m$ as it is often done with SAC \citep{SAC}, while employing the coefficients $\lbrace \, \theta_m \, \vert \, m = 1, \dots, M \, \rbrace$ in combination with the router $g$ in Eq. \eqref{eq:router}, with the learned parameters $\Theta$. In our evaluation campaign, per each environment, we test all the $N_{\text{train}} = 10$ policies on $N_{e} = 100$ independent episodes with maximum horizon $H=1000$ ($1000$ episodes in total), and we measure the average ER (avg. ER) achieved by the method. Numerical results on this are included in Tables \ref{tab:eval_interpretable}-\ref{tab:eval_non_interpretable}.

For comparison, we consider the results reported in \cite{nadizar2024naturally} for CGP and LGP, and the ones achieved by Metric-40 in \cite{akrour2021continuous} (where only 4 out of the 6 environments were tested, i.e., all except Reacher-v4 and Swimmer-v4), the best-performing policy trained in that work, characterized by a MoE architecture tuned with $40$ experts. The three methods are briefly discussed in the Related Work section. Moreover, we consider as non-linear competitors the classic version of Proximal Policy Search (PPO) \cite{schulman2017proximal}, and three neural architectures trained with SAC \cite{SAC}, characterized by different numbers of parameters. In particular, we consider SAC-M and SAC-S, previously described in the analysis of training performance, and we add a third, larger network, SAC-L. The architecture underlying SAC-L corresponds to the neural architecture employed in \cite{SAC}. For PPO, we consider the performance benchmarked in \cite{huang2022cleanrl}, while, for the SAC-based methods, we train and evaluate them with the same procedure described for \algname.

We can see in Table \ref{tab:eval_interpretable} that \algname consistently outperforms its competitors in five over six environments (all except Swimmer-v4). Additionally, Table \ref{tab:eval_non_interpretable} underlines that on average, our method's performance is closer to the one of the non-interpretable competitors, showing how \algname is performing consistently better than PPO (on five environments over six, all but Swimmer-v4), and on average close to the SAC-based approaches of comparable size, namely, SAC-M and SAC-S. 
Table \ref{tab:eval_non_interpretable} shows that SAC-L often achieves the best performance. This architecture is advantaged not only by exploiting non-linearity but also by the high number of parameters at its disposal (see Table \ref{tab:eval_non_interpretable}, last column), which, however, makes it computationally more expensive, both in training and in inference. 

As mentioned, Swimmer-v4 appears to be the most difficult environment for \algname, being the only one in which \algname does not outperform any of the interpretable methods, as well as PPO. 
To further comment on this, we can notice in Table \ref{tab:eval_non_interpretable} that the performance in this environment by all the SAC-based policies (including \algname) is weaker. 

\begin{figure*}[ht!]
\centering
\includegraphics[width=0.88\textwidth]{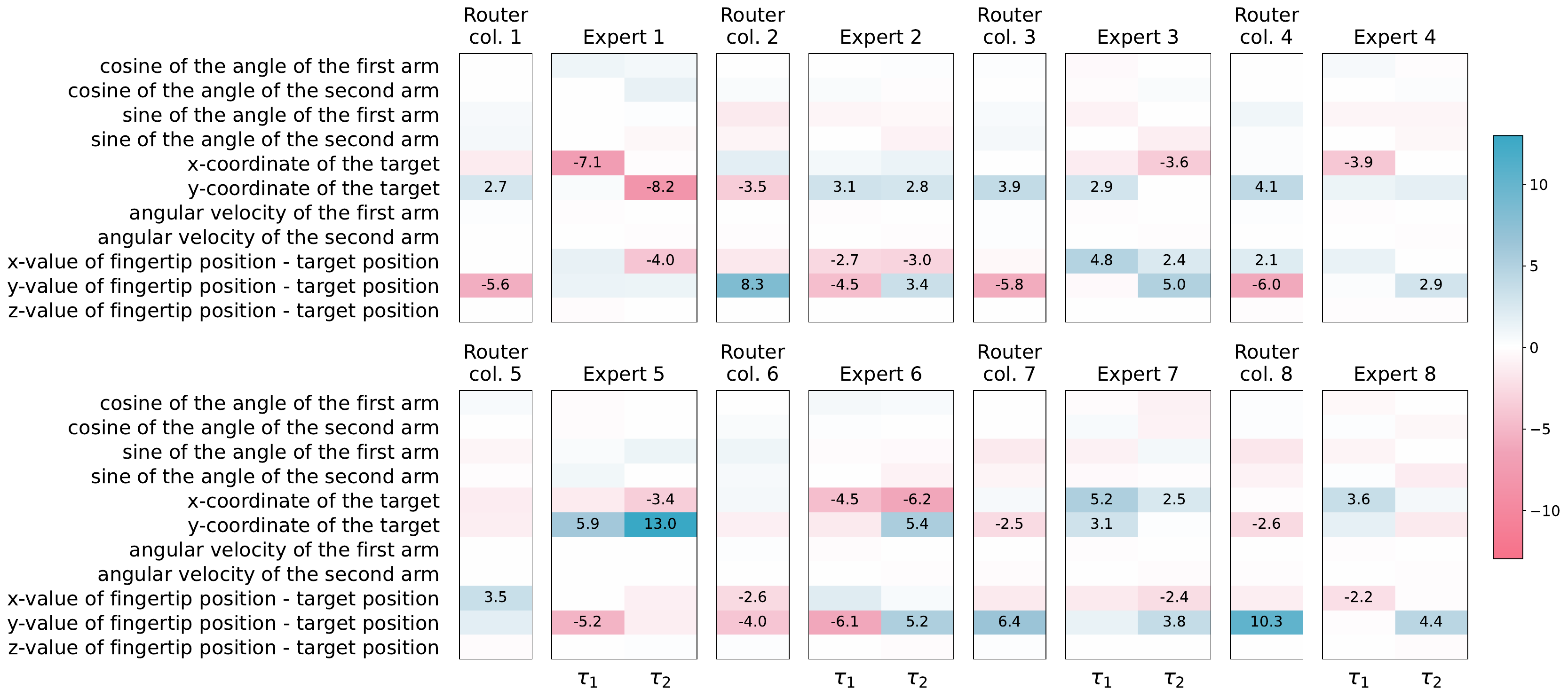}
\caption{\textbf{Reacher-v4}. Visualization of the weights for each expert and the corresponding column of the router's weight matrix.}
\label{fig:reacher_weights}
\end{figure*}

\subsection{Policy interpretation}
This section includes the interpretation of the best policy learned for the Reacher-v4 environment. 
We include interpretations for the other five environments in the Appendix. 
Figure \ref{fig:reacher_weights} presents a graphical representation of the weights of the MoE policy, including both the router and the experts. Details on both are included in the following, but we can already notice from the figure that 
the controller employs almost exclusively a small subset of variables:
\begin{itemize}
    \item coordinates of the target ($\rm T$): $\,\, x_{\rm T}, \, y_{\rm T}$
    \item coordinates' difference between the fingertip ($f$) and ${\rm T}$: $\,\,\Delta_x = x_f - x_{\rm T}, \,\,  \Delta_y = y_f - y_{\rm T}$
\end{itemize}
The two control variables, the torques applied to the first and second joint, are indicated as $\tau_1$ and $\tau_2$, respectively.
We indicate as $S_m$ the score of the $m$-th expert, i.e., the ``weight'' computed by the corresponding column of the router. 

\paragraph{Expert 1 ($S_1 \, \approx \, 2.7 \, y_{\rm T} - 5.6\, \Delta_y$)}
Expert 1 is called when its score $S_1$ is greater than all the other scores.
Its policy can be described as:
\begin{equation}
    \begin{cases}
        \tau_1 \, \approx \, x_{\rm T}, \\
        \tau_2 \, \approx \, y_{\rm T} - 4 \, \Delta_x .
    \end{cases}
\end{equation}
According to this, the first joint is strongly accelerated in a direction that is opposite to that of $x_{\rm T}$, while the second joint's control signal is composed of two terms, one that moves the joint in the opposite direction of $y_{\rm T}$, and the other that tries to minimize the distance on the x-axis between the fingertip and the target.

\paragraph{Expert 2 ($S_2 \, \approx \, -3.5 \, y_{\rm T} + 8.3 \, \Delta_y$)} It can be noted that Expert 2's score $S_2$ has opposite signs w.r.t. $S_1$, indicating that these two experts are likely working in opposite states. Its policy can be described as follows:
\begin{equation}
    \begin{cases}
        \tau_1 \, \approx \, 3.1 \, y_{\rm T} - 2.7 \, \Delta_x - 4.5 \, \Delta_y, \\
        \tau_2 \, \approx \, 2.8 \, y_{\rm T} + 3.0 \, \Delta_x - 3.4 \, \Delta_y,
    \end{cases}
\end{equation}
shows that the two joints' control signals have partially similar behaviors. Indeed, both of them have a dependency on $y_{\rm T}$, which can be seen as a feed-forward control scheme combined with a feed-back control scheme \cite{tao1994learning}.
This is suggested by the presence, in both controllers, of an amplified version of $y_{\rm T}$, and a negative dependency on $\Delta_y$ (please, note that in this environment $\Delta$ can be seen as the opposite of the control error).
We note, though, a strong difference between the two joints' controllers: while $\tau_1$ has a negative dependency on $\Delta_x$ (suggesting that it tries to reach the target also on the x-axis), $\tau_2$ shows the opposite. Interestingly, the magnitude of the two contributions is comparable, suggesting that these terms are used to balance the fingertip by simultaneously moving the two joints.

\paragraph{Expert 3 ($S_3 \, \approx \, 3.9 \, y_{\rm T} - 5.8 \, \Delta_y$)}
The score $S_3$ has similar coefficients to $S_1$, but with higher weight to $y_{\rm T}$, meaning that Expert 3 is preferred over Expert 1 when $y_{\rm T}$ has a high magnitude. The policy of this expert can be summarized as:
\begin{equation}
    \begin{cases}
        \tau_1 \, \approx \, 2.9 \, y_{\rm T} + 4.8 \, \Delta_x, \\
        \tau_2 \, \approx \, -3.6 \, x_{\rm T} + 2.4 \, \Delta_x + 5.0 \, \Delta_y.
    \end{cases}
\label{eq:exp3}
\end{equation}
Most of the terms in \eqref{eq:exp3} try to move away from the target (i.e., positive dependencies on $\Delta_x$ and $\Delta_y$, as well as a negative dependency on $x_{\rm T}$.
However, the positive dependency on $y_{\rm T}$ in $\tau_1$, shows an attempt to strike a balance between moving towards the desired $y_{\rm T}$ and moving away from $x_{\rm T}$. This likely builds momentum for reaching the target in the following steps, through the use of other experts. 

\paragraph{Expert 4 ($S_4 \, \approx \, 4.1 \, y_{\rm T} + 2.1 \, \Delta_x - 6.0 \,\Delta_y$)}
Expert 4 implements a simple policy, described by:
\begin{equation}
    \begin{cases}
        \tau_1 \, \approx \, -3.9 \, x_{\rm T}, \\
        \tau_2 \, \approx \, 2.9 \, \Delta_y.\\
    \end{cases}
\end{equation}
This policy tends to move away from the target.
In fact, in $\tau_1$ we have a negative dependency on $x_{\rm T}$, and in $\tau_2$ a positive one w.r.t. $\Delta_y = y_f - y_{\rm T}$, thus a positive weight pushing $y_f$ farther away from $y_{\rm T}$.

\paragraph{Expert 5 ($S_5 \, \approx \, 3.5 \, \Delta_x$)} This expert's policy is:
\begin{equation}
    \begin{cases}
        \tau_1 \, \approx \, 5.9 \, y_{\rm T} - 5.2 \, \Delta_y, \\
        \tau_2 \, \approx \, -3.4 \, x_{\rm T} + 13 \, y_{\rm T}.
    \end{cases}
\end{equation}
The torque $\tau_1$ aims for $y_f$ to reach $y_{\rm T}$ through combined feed-forward and feed-back control approaches, while $\tau_2$ pushes the fingertip to reach $y_{\rm T}$ only by using feed-forward control. Moreover, $\tau_2$ has a non-negligible negative dependency on $x_{\rm T}$, contributing to keeping the joint away from it.

\paragraph{Expert 6 ($S_6 \, \approx \, -2.6 \, \Delta_x - 4.0 \, \Delta_y$)}
Score $S_6$, suggests that Expert 6 is called when the difference between the fingertip's and the target's coordinates have large negative values. The policy works as follows:
\begin{equation}
    \begin{cases}
        \tau_1 \, \approx \, -4.5 \, x_{\rm T} - 6.1 \, \Delta_y, \\
        \tau_2 \, \approx \, -6.2 \, x_{\rm T} + 5.4 \, y_{\rm T} + 5.2 \, \Delta_y.
    \end{cases}
\end{equation}
While this policy is more intricate, we can interpret all the individual contributions, which will be combined at runtime (akin to the superposition principle in linear systems).
In $\tau_1$, we have a negative contribution from $x_{\rm T}$, pushing the joint away from it, and a negative contribution w.r.t. $\Delta_y$, moving the joint in such a way that the distance between the fingertip and the target (on the y-axis) is reduced.
On $\tau_2$ we have a negative dependency on $x_{\rm T}$, and a positive dependency on $y_{\rm T}$ (akin to feed-forward control), moving the joint in the same direction of $y_{\rm T}$, plus a positive dependency on $\Delta_y$, which makes the fingertip move away from the target on the y-axis.
While these two effects seem opposite, it is worth noticing that we can rework the equation as follows:
$\tau_2 \, \approx \, -6.2 \, x_{\rm T} + 5.4 \, y_{\rm T} + 5.2 \, \Delta_y = -6.2 \, x_{\rm T} + 5.4 \, y_{\rm T} + 5.2 \, y_f - 5.2 \, y_{\rm T} \, \approx \, -6.2 \, x_{\rm T} + 5.2 \, y_f$.
This revised equation can be easily interpreted as: $\tau_2$ tends to (i) move away from $x_{\rm T}$, and (ii) move towards the positive side of the y-axis.

\paragraph{Expert 7 ($S_7 \, \approx \, -2.5 \, y_{\rm T} + 6.4 \, \Delta_y$)}
Score $S_7$ suggests that Expert 7 is queried when the target is on the negative side of the y-axis, and the distance (also on the y-axis) has a large value (implying that the fingertip's y coordinate is larger than $y_{\rm T}$). The policy is:
\begin{equation}
    \begin{cases}
        \tau_1 \, \approx \, 5.2 \, x_{\rm T} + 3.1 \, y_{\rm T}, \\
        \tau_2 \, \approx \, 2.5 \, x_{\rm T} - 2.4 \, \Delta_x + 3.8. \Delta_y.
    \end{cases}
\end{equation}
We can see that $\tau_1$ uses a kind of feed-forward control to move towards both $x_{\rm T}$ and $y_{\rm T}$.
Instead, $\tau_2$, moves joint $2$ towards $x_{\rm T}$ (also exploiting $\Delta_x$), simultaneously moving it away from $y_{\rm T}$, through a positive contribution w.r.t. $\Delta_y$.

\paragraph{Expert 8 ($S_8 \, \approx \, -2.6 \, y_{\rm T} + 10.3 \, \Delta_y$)}
Score $S_8$, similarly to $S_7$, suggests that Expert 8 is called when $y_{\rm T} \leq 0$ and $y_f \geq y_{\rm T}$.
However, the larger weights suggest that this expert is called way more often than Expert 7. Its policy is:
\begin{equation}
    \begin{cases}
        \tau_1 \, \approx \,  3.6 \, x_{\rm T} - 2.2 \, \Delta_x ,\\
        \tau_2 \, \approx \,  4.4 \, \Delta_y.
    \end{cases}
\end{equation}
Here $\tau_1$ can be interpreted as simply performing combined feed-forward and feed-back control, while $\tau_2$ tends to move the joint farther from the target (on the y-axis).

\section{Conclusions}
In this paper, we introduced \algname, a novel approach for training sparsely activated and interpretable controllers using a \mbox{\emph{top}-$1$} Mixture-of-Experts architecture. By integrating interpretable shallow decision-makers, each specializing in different basic skills, and an interpretable router for task allocation, \algname strikes a balance between performance and interpretability. The evaluation of \algname presented in this work demonstrates its competitive performance, outperforming existing interpretable baselines and narrowing the performance gap with non-interpretable state-of-the-art methods. The transparency of \algname is also showcased in this work, through an in-depth interpretation. Additionally, by distilling DTs from the learned router, we supply an additional tool to facilitate the interpretation of the trained models, making \algname a compelling choice for scenarios where both high performance and interpretability are required. As future work, we plan to explore the potential of \algname in more complex environments and extend it to multi-agent scenarios, exploiting socially-inspired reward designs to achieve interpretable cooperative and coordinated AI policies.

\newpage
\section*{Acknowledgments}
We acknowledge ISCRA for awarding this project access to the LEONARDO supercomputer, owned by the EuroHPC Joint Undertaking, hosted by CINECA (Italy). 
This project is funded by the European Union. However, the views and opinions expressed are those of the author(s) only and do not necessarily reflect those of the European Union. Neither the European Union nor the granting authority can be held responsible for them.
Leonardo Lucio Custode and Giovanni Iacca acknowledge funding by the European Union (project no. 101071179). Laura Ferrarotti and Bruno Lepri acknowledge funding by the European Union’s Horizon Europe research and innovation program under grant agreement No. 101120237 (ELIAS) and under grant agreement No. 101120763 (TANGO).

\bibliography{smose}

\appendix

\section{Appendix}

\subsection{Computational setup}
All experiments were run on a single Intel Ice Lake CPU with 32 GB RAM on RedHat Enterprise Linux 8.6 operating system. In each environment, for each seed, the training took at most 6 hours to complete. The code was written in Python, using common packages for Reinforcement Learning, such as Gymnasium, PyTorch, and Stable-Baselines3. The exact versions used can be found in Table \ref{tab:packages}.

\begin{table}[ht!]
 \centering
 \caption{Package versions.}
 \label{tab:packages}
 \resizebox{0.5\columnwidth}{!}{
  \begin{tabular}{ll}
  \toprule
  \textbf{Package} & \textbf{Version} \\
  \midrule
  gymnasium & 0.29.1 \\
  matplotlib & 3.9.1 \\
  mujoco & 2.3.5 \\
  numpy & 2.0.1 \\
  pandas & 2.2.2 \\
  pip & 23.2.1 \\
  safetensors & 0.4.2 \\
  stable\_baselines3 & 2.3.2 \\
  torch & 2.1.2 \\
  \bottomrule
  \end{tabular}%
 }
\end{table}

\subsection{SAC parameters}
In Table \ref{tab:sac_params} we include the parameters used for Soft Actor-Critic (SAC) \cite{SAC}, for the training of \algname, and of the SAC-based competitors. These values come from the CleanRL project, without further tuning.

\begin{table}[ht!]
 \caption{SAC parameters}
 \label{tab:sac_params}
 \centering
 \resizebox{.8\columnwidth}{!}{
  \begin{tabular}{cll}
  \toprule
  & \textbf{Parameter}  & \textbf{Value} \\ \midrule
  \multirow{9}{*}{\rotatebox{90}{SAC}} & batch size  & 256  \\
  & buffer size & 1000000  \\
  & warm-up timesteps  & 10000 \\
  & actor learning rate & 3e-4  \\
  & critic learning rate  & 1e-3  \\
  & $\alpha$ learning rate & 1e-3  \\
  & discount factor $\gamma$  & 0.99  \\
  & target smoothing coefficient $\tau$ & 0.005 \\
  & target delay  & 2  \\
  & action log std range & (-5, 2)  \\ \bottomrule
  \end{tabular}%
 }
\end{table}

\subsection{Environments description}
This section contains a summarized description of the six environments employed as a benchmark in this paper. For additional details, the reader is referred to the online MuJoCo documentation \cite{MuJoCo}.

\begin{figure}[ht!]
\begin{subfigure}{.25\textwidth}
  \centering
  \includegraphics[width=.8\linewidth]{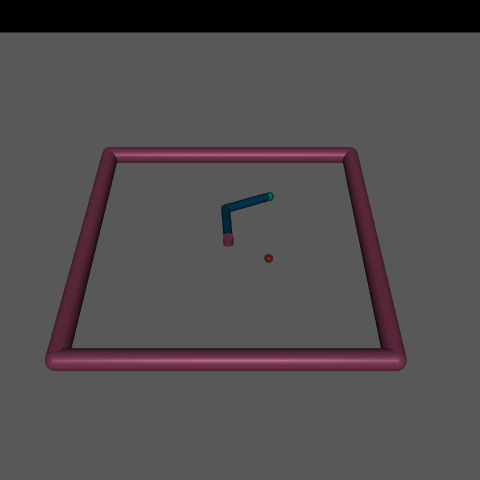}
  \caption{Reacher-v4}
  \label{fig:reacher}
\end{subfigure}%
\begin{subfigure}{.25\textwidth}
  \centering
  \includegraphics[width=.8\linewidth]{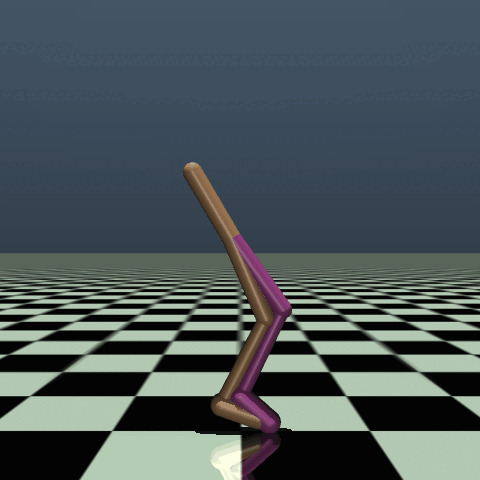}
  \caption{Walker2d-v4}
  \label{fig:walker}
\end{subfigure}
\caption{Visual representation of the Reacher-v4 and Walker2d-v4 environments.}
\end{figure}

\paragraph{Reacher-v4}(Fig. \ref{fig:reacher}) is a robotic arm with two joints. The objective in this environment is to maneuver the reacher's end effector, referred to as the fingertip, as close as possible to a target that is randomly positioned, by applying torques (ranging from -1 to +1) to the two joints. Observations include the cosine and sine of the angles of both arms, the coordinates of the target, the angular velocities of the arms, and the vector from the target to the reacher's fingertip. The assigned rewards penalize on one side, scenarios in which the reacher's fingertip is further away from the target, and on the other side, actions that are too large.

\paragraph{Walker2d-v4}(Fig. \ref{fig:walker}) is a two-dimensional, bipedal structure composed of seven primary body segments. These include a single torso at the top, from which the two legs diverge; two thighs located beneath the torso; two lower legs positioned below the thighs; and two feet attached to the lower legs, which support the entire structure. The objective is to achieve forward (rightward) motion by applying torques (ranging from -1 to +1) to the six joints connecting these seven body segments. Observations include the positional values of various body parts of the walker, followed by the corresponding velocities. A fixed reward is assigned for each timestep in which the walker is standing. Moreover, a positive reward is assigned if the walker walks forward, in the positive direction w.r.t. the x-coordinate. Finally, a penalty is assigned for actions that are too large.

\begin{figure}[ht!]
\begin{subfigure}{.25\textwidth}
  \centering
  \includegraphics[width=.8\linewidth]{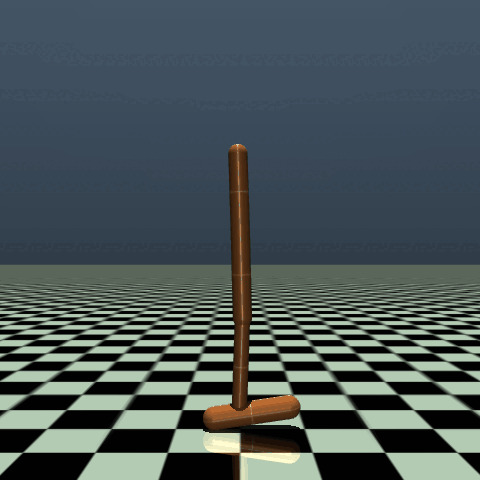}
  \caption{Hopper-v4}
  \label{fig:hopper}
\end{subfigure}%
\begin{subfigure}{.25\textwidth}
  \centering
  \includegraphics[width=.8\linewidth]{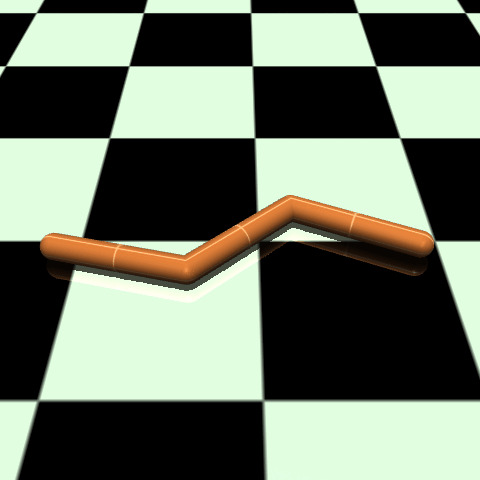}
  \caption{Swimmer-v4}
  \label{fig:swimmer}
\end{subfigure}
\caption{Visual representation of the Hopper-v4 and Swimmer-v4 environments.}
\end{figure}

\paragraph{Hopper-v4}(Fig. \ref{fig:hopper}) is a two-dimensional, single-legged structure composed of four primary body segments: the torso at the top, the thigh in the middle, the lower leg beneath the thigh, and a single foot supporting the entire body. The objective is to achieve forward (rightward) motion by generating hops through the application of torques (ranging from -1 to +1) at the three joints connecting these four segments. The structure of the observations and rewards is similar to the ones described for Walker2d-v4.

\paragraph{Swimmer-v4}(Fig. \ref{fig:swimmer}) consists of three segments connected by two articulation joints (called rotors), each of which connects exactly two links to form a linear chain. The swimmer is suspended in a two-dimensional pool and always starts from a similar position (with minor deviations drawn from a uniform distribution). The objective is to move as quickly as possible to the right by applying torques  (ranging from -1 to +1) to the joints and utilizing fluid friction. The structure of the observations is similar to the ones described for Walker2d-v4. A positive reward is assigned if the swimmer moves forward, while a penalty is associated with excessively large actions.

\begin{figure}[ht!]
\begin{subfigure}{.25\textwidth}
  \centering
  \includegraphics[width=.8\linewidth]{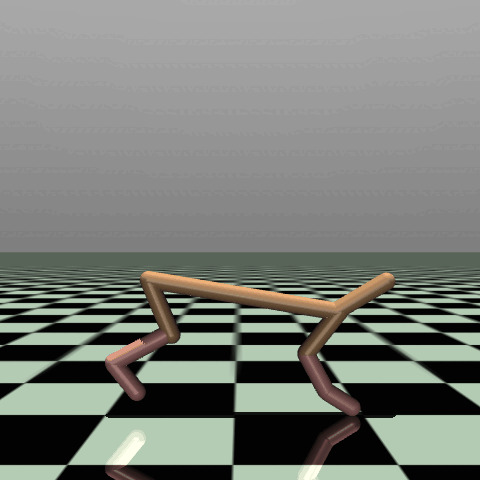}
  \caption{HalfCheetah-v4}
  \label{fig:halfcheetah}
\end{subfigure}%
\begin{subfigure}{.25\textwidth}
  \centering
  \includegraphics[width=.8\linewidth]{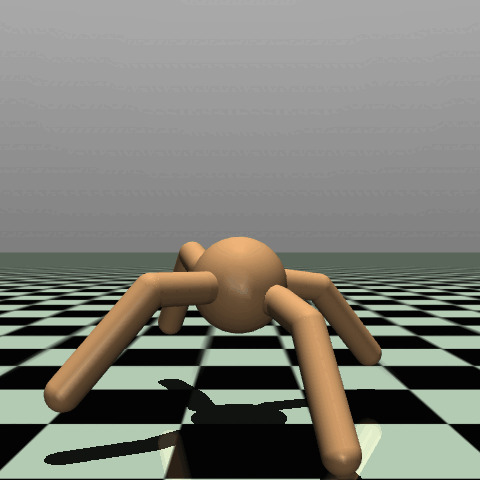}
  \caption{Ant-v4}
  \label{fig:ant}
\end{subfigure}
\caption{Visual representation of the HalfCheetah-v4 and Ant-v4 environments.}
\end{figure}

\paragraph{HalfCheetah-v4}(Fig. \ref{fig:halfcheetah}) is a 2-dimensional robot composed of 9 body parts connected by 8 joints, including two paws. The objective is to apply torques (ranging from -1 to +1) to the joints to make the cheetah run forward (to the right) as quickly as possible. Positive rewards are given based on the distance traveled forward, while negative rewards are given for moving backwards. The cheetah's torso and head are fixed, and torques can only be applied to the 6 joints located at the front and back thighs (connected to the torso), shins (connected to the thighs), and feet (connected to the shins).

\paragraph{Ant-v4}(Fig. \ref{fig:ant}) is a 3-dimensional robot composed of a central torso, which can rotate freely, and four legs, each consisting of two segments. The objective is to coordinate the movement of the four legs to achieve forward (rightward) motion by applying torques (ranging from -1 to +1) to the eight hinges that connect the leg segments to each other and to the torso. In total, the robot has nine body parts and eight hinges.

\subsection{Number of experts ablation}
\label{app:ablation_M}
The number of experts employed by \algname within the MoE architecture is represented by the $M$ parameter. Such parameter has been empirically tuned considering five different values ranging from $3$ to $32$, to balance agents' performance (to be maximized) and the size of the mixture of experts (to be minimized). We found that lowering the number of experts below $M=8$ decreases performance, while increasing the number of experts above $M=8$ increases the required interpretation effort. These considerations are substantiated by the results of the ablation study included in Table \ref{tab:num_experts}.

\begin{table}[ht!]
 \centering
 \caption{\textbf{Ablation on the number of experts.}}
 \label{tab:num_experts}
 \resizebox{\columnwidth}{!}{
  \begin{tabular}{lcc}
  \toprule
  \textbf{Environment} & \textbf{N. experts $M$} & \multicolumn{1}{c}{\textbf{avg. ER}}\\
  \midrule
  \multirow{5}{*}{Walker2d-v4} & 3 & $3920.91 \phantom{0}\pm\phantom{00}14.42$\\
                               & 5 & $4209.33 \phantom{0}\pm\phantom{00}10.71$\\
                               & 8 & $4224.29 \phantom{0}\pm\phantom{00}25.96$\\
                               & 16& $4459.63 \phantom{0}\pm\phantom{00}20.69$\\
                               & 32& $4177.62 \phantom{0}\pm\phantom{00}29.81$\\
  \midrule
  \multirow{5}{*}{Hopper-v4}   & 3 & $2358.15 \phantom{0}\pm\phantom{00}16.09$\\
                               & 5 & $2521.34 \phantom{0}\pm\phantom{0}192.76$\\
                               & 8 & $2816.08 \phantom{0}\pm\phantom{0}445.57$\\
                               & 16& $3213.00 \phantom{0}\pm\phantom{0}210.94$\\
                               & 32& $3329.36 \phantom{0}\pm\phantom{0}106.94$\\
  \midrule
  \multirow{5}{*}{Ant-v4} & 3 & $3217.14 \phantom{0}\pm\phantom{0}526.53$\\
                          & 5 & $3177.08 \phantom{0}\pm\phantom{0}775.01$\\
                          & 8 & $3245.43 \phantom{0}\pm\phantom{0}380.93$\\
                          & 16& $3243.08 \phantom{0}\pm\phantom{0}565.66$\\
                          & 32& $3522.76 \phantom{0}\pm\phantom{0}506.69$\\
  \midrule
  \multirow{5}{*}{HalfCheetah-v4} & 3 & $6349.63 \phantom{0}\pm\phantom{00}67.60$\\
                                  & 5 & $7073.59 \phantom{0}\pm\phantom{00}94.46$\\
                                  & 8 & $7310.17 \phantom{0}\pm\phantom{0}131.57$\\
                                  & 16& $7925.09 \phantom{0}\pm\phantom{00}83.27$\\
                                  & 32& $7970.24 \phantom{0}\pm\phantom{00}97.02$\\
  \midrule
  \multirow{5}{*}{Reacher-v4} & 3 & $\;\;\,-6.57 \phantom{0}\pm\phantom{000}3.51$\\
                              & 5 & $\;\;\,-6.23 \phantom{0}\pm\phantom{000}3.46$\\
                              & 8 & $\;\;\,-5.49 \phantom{0}\pm\phantom{000}2.32$\\
                              & 16& $\;\;\,-4.94 \phantom{0}\pm\phantom{000}2.05$\\
                              & 32& $\;\;\,-4.95 \phantom{0}\pm\phantom{000}1.92$\\
  \midrule
  \multirow{5}{*}{Swimmer-v4} & 3 & $\phantom{00}44.45 \phantom{0}\pm\phantom{000}1.37$\\
                              & 5 & $\phantom{00}45.72 \phantom{0}\pm\phantom{000}1.92$\\
                              & 8 & $\phantom{00}45.40 \phantom{0}\pm\phantom{000}1.62$\\
                              & 16& $\phantom{00}47.41 \phantom{0}\pm\phantom{000}1.29$\\
                              & 32& $\phantom{00}46.66 \phantom{0}\pm\phantom{000}1.38$\\
  \bottomrule
  \end{tabular}%
 }
\end{table}

\subsection{Decision Trees as a support for interpretation}

Once the policy training is complete, thanks to the \algname architecture we obtain a policy that is inherently interpretable. To further enhance the readability of our policy and improve the ease of interpretation, we distill a multi-class classifier 
from the learned router $g(\, s \, \vert \, \Theta \, )$, considering $M$ classes, one per each of the $M$ experts $\pi_m$. This classifier can be used as an extremely readable support for the interpretation of the learned policy, that remains anyhow directly interpretable thanks to the shallow nature of its components.

With the goal of enhancing easiness of interpretation, for the distilled classifier we exploit DTs. A DT is a hierarchical model used for decision-making, where data is recursively split based on feature values, leading to easily interpretable decisions; this structure makes DTs particularly effective for interpretability, as they provide clear, visual representations of the decision-making process \cite{good2023feature}. In order to achieve easily explainable trees, it is necessary to consider architectures with limited depth, potentially decrementing their capability of approximating the original continuously parameterized router $g(\, s \, \vert \, \Theta \, )$. To achieve a trade-off between depth and expressive power, we split the multi-class classification problem into $M$ simpler binary classification tasks. We distill $M$ DTs $\lbrace \, \text{DT}_m \, \rbrace$, each one associated with a specific expert policy $\pi_m$, evaluating, given a state $s$ in input, if the control task in such state should be assigned to the $m$-th expert or not, i.e.:
\begin{equation*}
 \begin{cases}
  \text{DT}_m(\Theta^m \cdot s) = 1 & \text{if expert $m$ is assigned to state $s$} \\
  \text{DT}_m(\Theta^m \cdot s) = 0 & \text{otherwise.}
 \end{cases} 
\end{equation*}
Here $\Theta^m$ indicates the $m$-th column of matrix $\Theta$. To obtain the DTs, we use the data collected in the replay buffer during the training of the interpretable MoE policy. We evaluate the states within the replay buffer and label them using the trained router $g(\, s \, \vert \, \Theta \, )$, assigning each state to one of the $M$ experts. We then create $M$ datasets, as copies of the labeled dataset: considering the $m$-th dataset, we relabel the data by assigning label $m$ as 1 and setting all other labels to 0. After balancing the data with weights that are inversely proportional to class frequency, we train the $m$-th DT (denoted as $\text{DT}_m$) in a supervised manner, using the CART algorithm \cite{timofeev2004classification}, with fixed maximum depth $d$.

\begin{figure}[!ht]
\centering
\includegraphics[width=0.99\columnwidth]{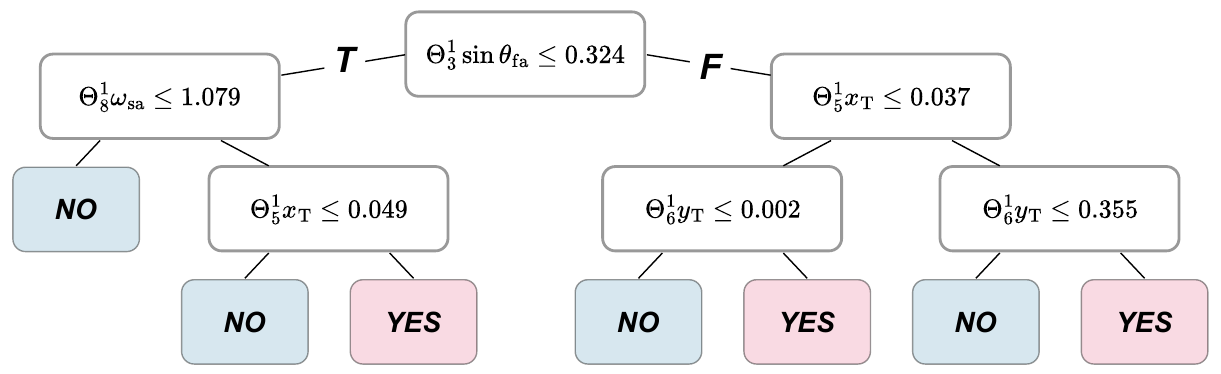}
\caption{\textbf{Reacher-v4}. Decision tree for Expert 1.}
\label{fig:reacher_dt1}
\end{figure}

\begin{figure}[!ht]
\centering
\includegraphics[width=0.99\columnwidth]{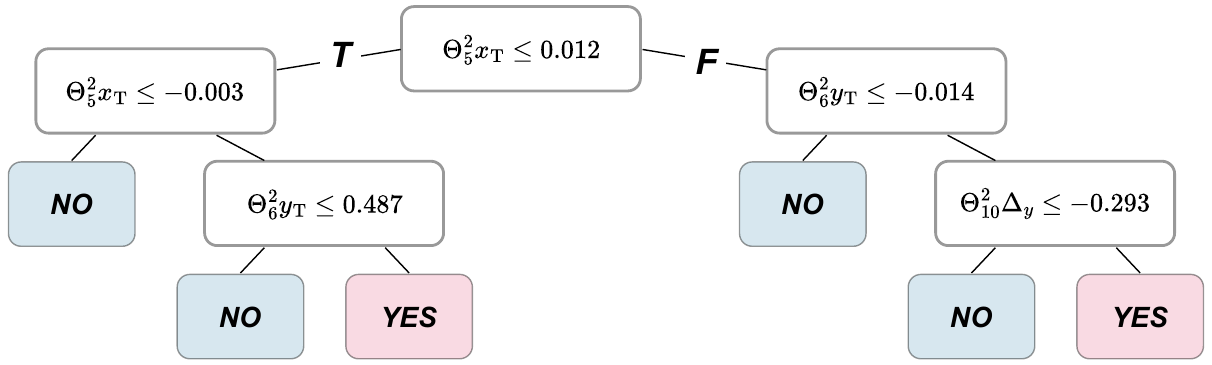}
\caption{\textbf{Reacher-v4}. Decision tree for Expert 2.}
\label{fig:reacher_dt2}
\end{figure}

\begin{figure}[!ht]
\centering
\includegraphics[width=0.99\columnwidth]{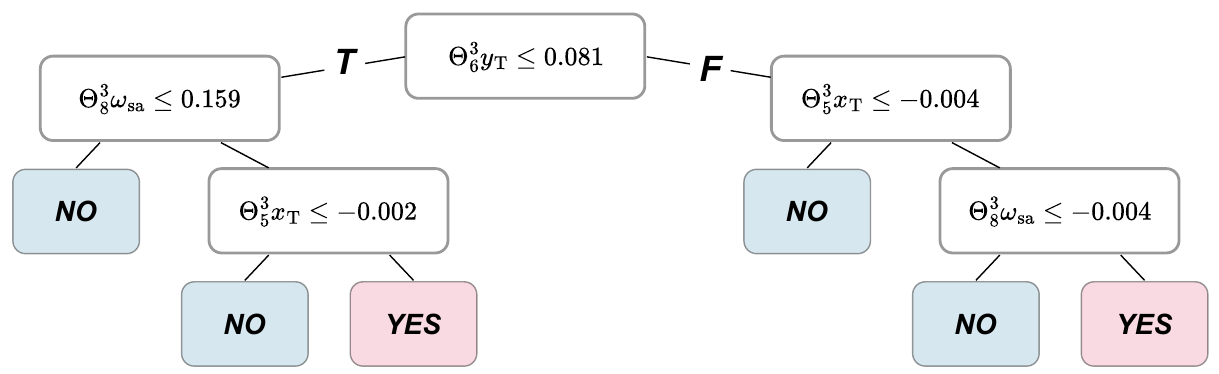}
\caption{\textbf{Reacher-v4}. Decision tree for Expert 3.}
\label{fig:reacher_dt3}
\end{figure}

\begin{figure}[!ht]
\centering
\includegraphics[width=0.99\columnwidth]{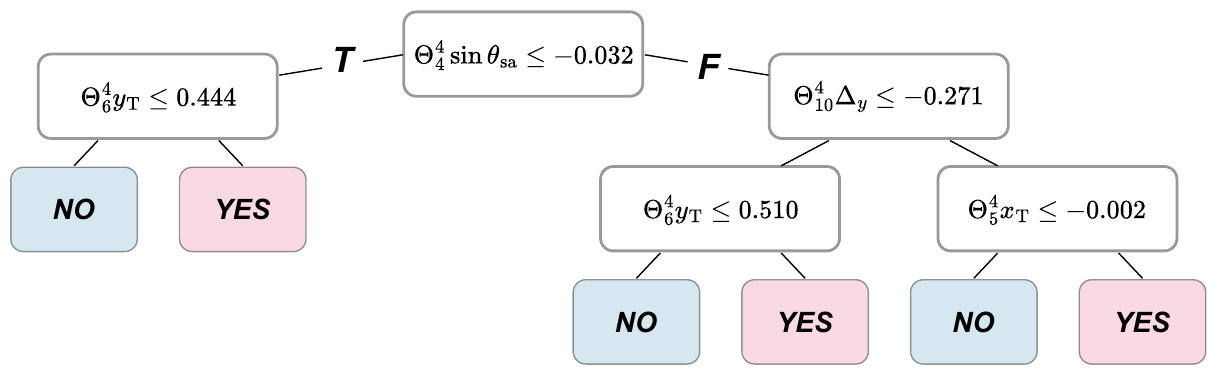}
\caption{\textbf{Reacher-v4}. Decision tree for Expert 4.}
\label{fig:reacher_dt4}
\end{figure}

\begin{figure}[!ht]
\centering
\includegraphics[width=0.99\columnwidth]{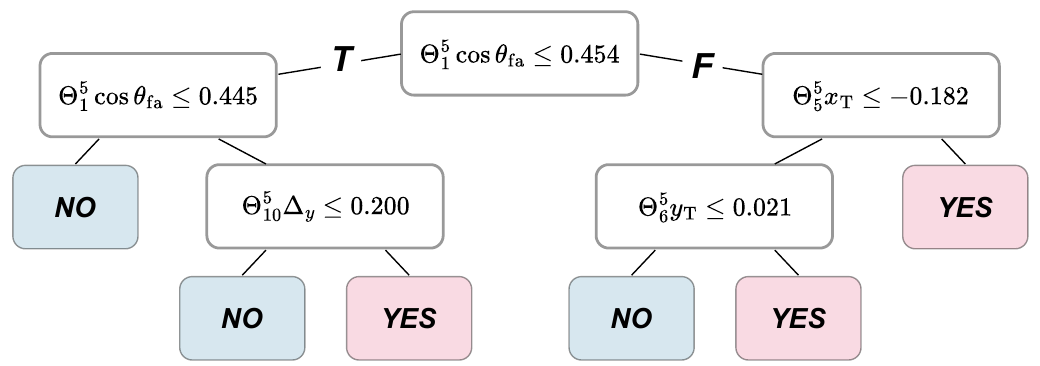}
\caption{\textbf{Reacher-v4}. Decision tree for Expert 5.}
\label{fig:reacher_dt5}
\end{figure}

\begin{figure}[!ht]
\centering
\includegraphics[width=0.99\columnwidth]{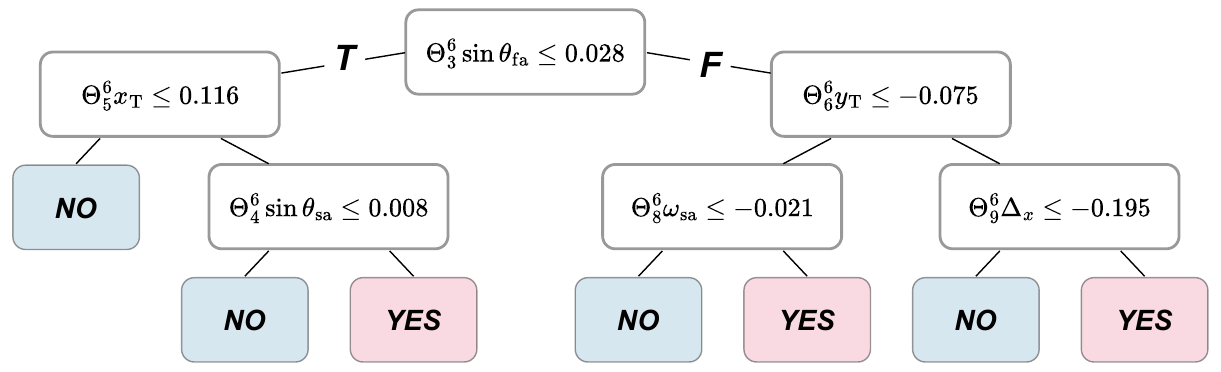}
\caption{\textbf{Reacher-v4}. Decision tree for Expert 6.}
\label{fig:reacher_dt6}
\end{figure}

\begin{figure}[!ht]
\centering
\includegraphics[width=0.99\columnwidth]{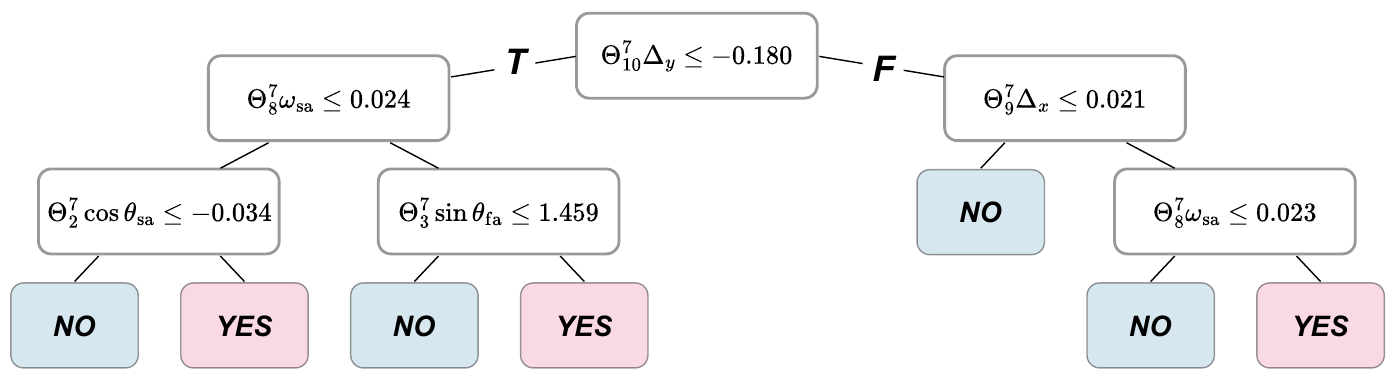}
\caption{\textbf{Reacher-v4}. Decision tree for Expert 7.}
\label{fig:reacher_dt7}
\end{figure}

\begin{figure}[!ht]
\centering
\includegraphics[width=0.99\columnwidth]{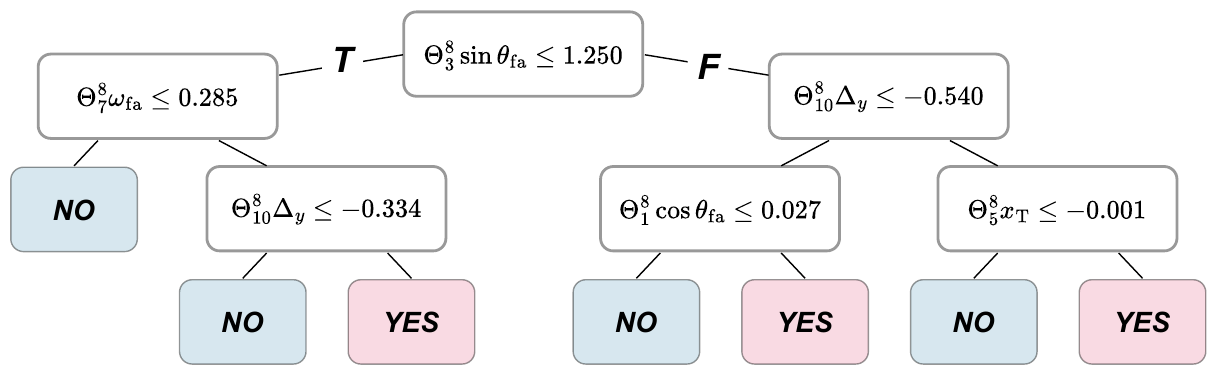}
\caption{\textbf{Reacher-v4}. Decision tree for Expert 8.}
\label{fig:reacher_dt8}
\end{figure}

In this work, we propose to distill the trees in a simple supervised setting, but more sophisticated imitation learning solutions, e.g., inspired by \cite{ross2011reductionimitationlearningstructured, bastani2019verifiablereinforcementlearningpolicy}, could be obtained.
In the following, we include the DTs distilled from the router of Reacher-v4, as an example. Figures \ref{fig:reacher_dt1}-
\ref{fig:reacher_dt8} show the distilled trees of depth $d=3$ associated with $\text{DT}_1$, $\text{DT}_2$, $\text{DT}_3$, $\text{DT}_4$, $\text{DT}_5$, $\text{DT}_6$, $\text{DT}_7$, and $\text{DT}_8$, respectively. In the trees, $\theta_{\rm fa}$ and $\theta_{\rm sa}$ indicate, respectively, the angle of the first and second arm; $\omega_{\rm fa}$ and $\omega_{\rm sa}$ indicate, respectively, the angular velocity of the first and second arm.

\subsection{Additional policy interpretations}
In this section, we will interpret the best solutions obtained for the Walker2d-v4, Hopper-v4, Swimmer-v4, HalfCheetah-v4, and Ant-v4 environments.

\subsubsection{Walker2d-v4}
\begin{figure*}[ht!]
\centering
\includegraphics[width=1\textwidth]{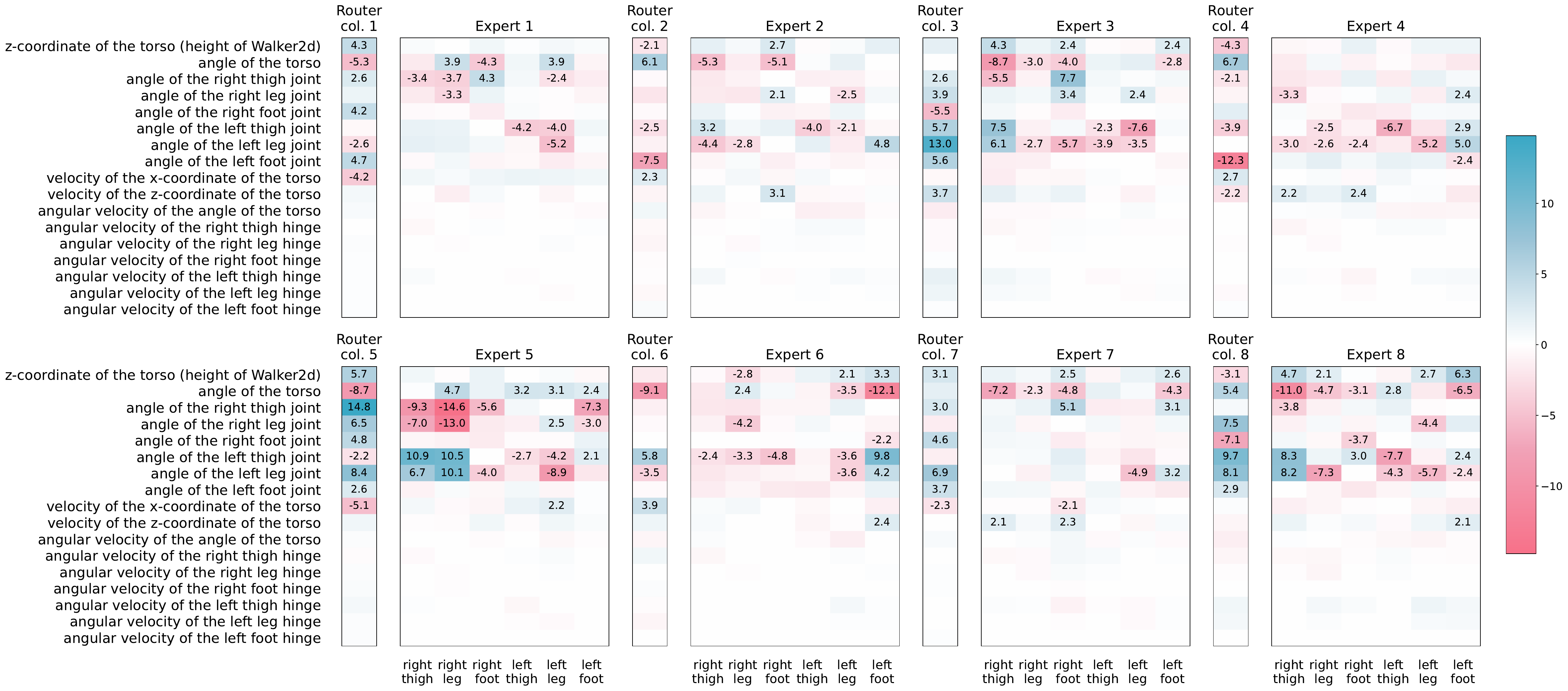}
\caption{\textbf{Walker2d-v4}. Visual representation of the learned weights for each expert and of the corresponding column of the router’s weight matrix.}
\label{fig:walker_weights}
\end{figure*}
In this environment, the policy makes use of the following variables:
\begin{itemize}
    \item z-coordinate of the torso: $z_{t}$
    \item angle of the torso: $\theta_{t}$
    \item angle of right thigh: $\theta_{rt}$
    \item angle of right leg: $\theta_{rl}$
    \item angle of right foot: $\theta_{rf}$
    \item angle of left thigh: $\theta_{lt}$
    \item angle of left leg: $\theta_{ll}$
    \item angle of left foot: $\theta_{lf}$
    \item x-velocity of the torso: $v_{x}$
    \item z-velocity of the torso: $v_{z}$
    \item angular velocity of the torso: $\omega_{t}$
    \item angular velocity of right thigh: $\omega_{rt}$
    \item angular velocity of right leg: $\omega_{rl}$
    \item angular velocity of right foot: $\omega_{rf}$
    \item angular velocity of left thigh: $\omega_{lt}$
    \item angular velocity of left leg: $\omega_{ll}$
    \item angular velocity of left foot: $\omega_{lf}$
\end{itemize}

to control the following variables:
\begin{itemize}
    \item right thigh's torque: $\tau_{rt}$
    \item right leg's torque: $\tau_{rl}$
    \item right foot's torque: $\tau_{rf}$
    \item left thigh's torque: $\tau_{lt}$
    \item left leg's torque: $\tau_{ll}$
    \item left foot's torque: $\tau_{lf}$
\end{itemize}

\paragraph{Expert 1 ($S_1 \approx 4.3 z_t - 5.3 \theta_t + 2.6 \theta_{rt} + 4.2 \theta_{rf} - 2.6 \theta_{ll} + 4.7 \theta_{lf} - 4.2 v_x$)}
This expert's scoring function suggests that it is queried when the z-coordinate of the torso is large and its angle is negative, together with having a positive angle at the right thigh, the right leg, and the left foot, while having negative values for the left leg's angle and x-velocity.
This indicates that the walker has its right thigh and foot raised while having its left leg behind (i.e., it has just taken a step with the left foot).
Moreover, the negative weight assigned to the x-velocity by this scoring function suggests that this expert is much more preferred when the torso is moving towards the left, indicating a potential unbalance.

The policy of this expert is the following:
\begin{equation}
\begin{cases}
    \tau_{rt} \approx -3.4 \theta_{rt} \\
    \tau_{rl} \approx 3.9 \theta_{t} - 3.7 \theta_{rt} - 3.3 \theta_{rl} \\
    \tau_{rf} \approx  -4.3 \theta_t + 4.3 \theta_{rt}\\
    \tau_{lt} \approx -4.2 \theta_{lt} \\
    \tau_{ll} \approx 3.9 \theta_t - 2.4 \theta_{rt} - 4.0 \theta_{lt} - 5.2 \theta_{ll}\\
    \tau_{lf} \approx 0 \\
\end{cases}
\end{equation}

The torque applied to the right thigh is negatively proportional to that thigh's angle, indicating that this expert seeks to bring the angle to 0 (i.e., moves it ``backward'').
Similarly, $\tau_{rl}$ applies a similar mechanism to its own joint. However, it also uses two additional contributions: one based on the torso's angle, and another one based on the right thigh's angle. This indicates that this leg tends to move in accordance with its thigh, while also taking into account the torso.
Regarding the right foot, its policy is significantly different. In fact, it has two terms (of equal magnitude, interestingly): one for the torso, and the other one for the right thigh. By reworking its equation, we observe that it simply moves in the same direction as the difference between those two angles:
\begin{equation}
    \tau_{rf} \approx 4.3 (\theta_{rt} - \theta_t)
\end{equation}
The torque applied to the left thigh is very similar to that of the right thigh but applied to the left joint. In contrast,
$\tau_{ll}$ consists of three contributions: one based on the torso's angle (interestingly enough, with the exact same coefficient as $\tau_{rl}$), one opposed to the right thigh's angle, another opposed to the left thigh's angle, and the last opposed to its own angle (which indicates that this torque, besides coordinating with the other joints, also aims to bring its own joint's angle close to $0$).

\paragraph{Expert 2 ($S_2 \approx -2.1 z_t + 6.1 \theta_t - 2. \theta_{lt} - 7.5 \theta_{lf} + 2.3 v_x$)}
This expert is likely queried when the z-coordinate of the torso is negative, the torso's angle is positive, the left thigh's and foot's angles are negative and the x-velocity is positive.

Its policy is:
\begin{equation}
\begin{cases}
    \tau_{rt} \approx -5.3 \theta_t + 3.2 \theta_{lt} - 4.4 \theta_{ll}\\
    \tau_{rl} \approx -2.8 \theta_{ll}\\
    \tau_{rf} \approx 2.7 z_t - 5.1 \theta_t + 2.1 \theta_{rl} + 3.1 v_z\\
    \tau_{lt} \approx -4.0 \theta_{lt}\\
    \tau_{ll} \approx -2.5 \theta_{rl} - 2.1 \theta_{lt}\\
    \tau_{lf} \approx 4.8 \theta_{ll}\\
\end{cases}
\end{equation}
We observe that the right thigh's torque tends to balance the torso's angle, while it applies a torque based on both the left thigh's and the left leg's angles, to coordinate the two legs' movements.
The torque applied to the right leg, interestingly, only depends on the left leg's angle.
The right foot's torque, instead, is a bit more complex.
In fact, it depends positively on the z-coordinate, the right leg's angle, and the z-velocity; while it has a negative dependency on the torso's angle.
Regarding the left part, $\tau_{lt}$ is only dependent on the angle of the same joint, trying to reduce its magnitude.
The torque applied to the right leg, instead, is both based on the right leg and the left thigh, suggesting that its role is to (1) balance the thigh's movement, and (2) coordinate with the left leg.
Finally, the torque applied to the right foot tries to preserve the momentum of the joint itself.

\paragraph{Expert 3 ($S_3 \approx 2.6 \theta_{rt} + 3.9 \theta_{rl} - 5.5 \theta_{rf} + 5.7 \theta_{lt} + 13.0 \theta_{ll} + 5.6 \theta_{lf} + 3.7 v_z$)}
The scoring function of this expert suggests that it is queried when all the joints' angles are positive, except for the torso's angle (which has a negligible weight) and the right foot's angle, which has a negative weight.
Moreover, it is dependent on the z-velocity being positive, suggesting that this expert is likely responsible for handling the time between one step and another.

Its policy is:
\begin{equation}
\begin{cases}
    \tau_{rt} \approx 4.3 z_t - 8.7 \theta_t - 5.5 \theta_{rt} + 7.5 \theta_{lt} + 6.1 \theta_{ll}\\
    \tau_{rl} \approx -3.0 \theta_t - 2.7 \theta_{ll}\\
    \tau_{rf} \approx 2.4 z_t - 4.0 \theta_t + 7.7 \theta_{rt} + 3.4 \theta_{rl} - 5.7 \theta_{ll}\\
    \tau_{lt} \approx -2.3 \theta_{lt} - 3.9 \theta_{ll}\\
    \tau_{ll} \approx 2.4 \theta_{rl} - 7.6 \theta_{lt} - 3.5 \theta_{ll}\\
    \tau_{lf} \approx 2.4 z_t - \theta_t\\
\end{cases}
\end{equation}
The right thigh's torque has three main roles: (1) it moves the leg forward to prepare the walker for the next step when its z-coordinate is large; (2) it balances the torso's and the thigh's angles; and (3) it coordinates the right thigh with the other leg.
The torque applied to the right leg is much simpler.
It has a term that is dependent on the torso's angle, and another term that depends on the left leg, likely to coordinate the two legs' movements.
The torque applied to the right foot has the following positive contributions (1) the z-coordinate (similarly to $\tau_{rt}$), (2) the right thigh's angle, and (3) the right leg's angle; while it depends negatively on the torso's angle and the left leg's angle.
Regarding the left part, the left thigh's torque tries to balance its own joint, while also having a negative dependency on the left leg's angle.
The left leg's torque has a behavior that is similar to that of the left thigh, but it also depends (positively) on the right foot's angle.
Finally, the left foot's torque only depends on the z-coordinate (positively) and the torso's angle (negatively).

It is interesting to note that, in this expert, most of the torques have a non-negligible dependency on the left leg's angle, which suggests that this joint is central to the policy of this expert.

\paragraph{Expert 4 ($S_4 \approx -4.3 z_t + 6.7 \theta_t - 2.1 \theta_{rt} - 3.9 \theta_{lt} - 12.3 \theta_{lf} + 2.7 v_x - 2.2 v_z$)}
This expert's scoring function is very similar to that of Expert 2. However, it has two main additional terms one depending on the right thigh's angle, and the other depending on the z-velocity. This suggests that this expert is preferred over Expert 2 when the right thigh's angle is negative, and the z-velocity is negative.

Its policy is:
\begin{equation}
\begin{cases}
    \tau_{rt} \approx -3.3 \theta_{rl} - 3.0 \theta_{ll} + 2.2 v_z\\
    \tau_{rl} \approx -2.5 \theta_{lt} - 2.6 \theta_{ll}\\
    \tau_{rf} \approx -2.4 \theta_{ll} + 2.4 z_t\\
    \tau_{lt} \approx -6.7 \theta_{lt}\\
    \tau_{ll} \approx -5.2 \theta_{ll}\\
    \tau_{lf} \approx 2.4 \theta_{rl} + 2.9 \theta_{lt} + 5.0 \theta_{ll} - 2.4 \theta_{lf}\\
\end{cases}
\end{equation}
Also in this case, we observe that the policy significantly depends on the left leg's angle, and, in general, on the left part of the robot.
The right thigh's torque is dependent on both leg's angles, likely to balance the two legs' movements; and the z-velocity, to improve the momentum of the whole robot.
The right leg, instead, moves in the opposite direction of the joints of the left thigh and leg.
The right foot's torque also depends on the left leg, while also having a positive contribution based on the z-velocity, likely to increase the acceleration generated by the movement of the foot on the ground.
The left thigh and the left leg only seek to balance their position.
Finally, the left foot moves according to the right leg's angle, and the left thigh's and leg's angles, while having a negative contribution w.r.t. its own angle.

\paragraph{Expert 5 ($S_5 \approx 5.7 z_t - 8.7 \theta_t + 14.8 \theta_{rt} + 6.5 \theta_{rl} + 4.8 \theta_{rf} - 2.2 \theta_{lt} + 8.4 \theta_{ll} + 2.6 \theta_{lf} - 5.1 v_x$)}
The difference between the scoring function of Expert 5 w.r.t. Expert 1 mainly lies in the different signs for the left leg's joint, and the two dependencies on the right leg's angle and the left thigh's angles.

The policy of this expert is a bit more complex:
\begin{equation}
\begin{cases}
    \tau_{rt} \approx -9.3 \theta_{rt} - 7.0 \theta_{rl} + 10.9 \theta_{lt} + 6.7 \theta_{ll}\\
    \tau_{rl} \approx 4.7 \theta_t - 14.6 \theta_{rt} - 13.0 \theta_{rl} + 10.5 \theta_{lt} + 10.1 \theta_{ll}\\
    \tau_{rf} \approx -5.6 \theta_{rt} - 4.0 \theta_{ll}\\
    \tau_{lt} \approx 3.2 \theta_t - 2.7 \theta_{lt}\\
    \tau_{ll} \approx 3.1 \theta_t + 2.5 \theta_{rl} - 4.2 \theta_{lt} - 8.9 \theta_{ll} + 2.2 v_x\\
    \tau_{lf} \approx 2.4 \theta_t - 7.3 \theta_{rt} - 3.0 \theta_{rl} + 2.1 \theta_{lt}\\
\end{cases}
\end{equation}
The torque applied to the right thigh depends negatively on the joint itself and the right leg, while it tends to move according to the left thigh and the left leg.
The torque applied to the right leg has a behavior similar to that applied to the right thigh, with the difference of an additional term depending on the angle of the torso.
The right foot's torque is much simpler: it only makes use of the right thigh's angle and the left leg's angle, both multiplied by a negative weight.
The left thigh's torque is computed by using a contribution that is proportional to the torso's angle and a negatively weighted version of the left thigh's angle.
The left leg's torque uses the same variables as the left thigh, plus two positive contributions (from the right leg's angle and the x-velocity), and a negative contribution (from its own joint's angle).
Finally, the left foot's torque depends positively on the torso's angle and the left thigh's angle, and negatively on the right thigh's and right leg's angles.
 
\paragraph{Expert 6 ($S_6 \approx -9.1 \theta_t + 5.8 \theta_{lt} - 3.5 \theta_{ll} + 3.9 v_x $)}
This expert is likely called when the torso has a negative angle, the left thigh's angle is positive, the left leg's angle is negative, and the x-velocity is positive.

This expert's policy is:
\begin{equation}
\begin{cases}
    \tau_{rt} \approx -2.4 \theta_{lt}\\
    \tau_{rl} \approx -2.8 z_t + 2.4 \theta_t - 4.2 \theta_{rl} - 3.3 \theta_{lt}\\
    \tau_{rf} \approx -4.8 \theta_{lt}\\
    \tau_{lt} \approx 0\\
    \tau_{ll} \approx 2.1 z_t - 3.5 \theta_t - 3.6 \theta_{lt} - \theta_{ll}\\
    \tau_{lf} \approx 3.3 z_t - 12.1 \theta_t - 2.2 \theta_{rt} + 9.8 \theta_{lt} + 4.2 \theta_{ll} + 2.4 v_z\\
\end{cases}
\end{equation}
We observe that this policy significantly depends on the left thigh's angle.
In fact, $\tau_{rt}$ and $\tau_{rf}$ only depend on it.
$\tau_{rl}$ depends negatively on the z-coordinate, the right leg's joint, and the left thigh's joint; while it depends positively on the torso's angle.
The left thigh does not get any significant torque.
The left leg's torque, instead, depends positively on the z-coordinate and negatively on the torso's, the left leg's, and the left foot's angles.
Finally, the left foot's torque depends positively on the z-coordinate and the left thigh's, left leg's angles, and z-velocity; while depending negatively on the torso's angle and the right foot's angle.

\paragraph{Expert 7 ($S_7 \approx 3.1 z_t + 3.0 \theta_{rt} + 4.6 \theta_{rf} + 6.9 \theta_{ll} + 3.7 \theta_{lf} - 2.3 v_x$)}
The scoring function of this policy indicates that this expert is likely to be called when the x-velocity is negative and the z-coordinate, the right thigh's angle, the right foot's angle, the left leg's angle, and the left foot's angle are positive.

This expert's policy can be summarized as:
\begin{equation}
\begin{cases}
    \tau_{rt} \approx -7.2 \theta_t + 2.1 v_z\\
    \tau_{rl} \approx -2.3 \theta_t\\
    \tau_{rf} \approx 2.5 z_t - 4.8 \theta_t + 5.1 \theta_{rt} - 2.1 v_x + 2.3 v_z\\
    \tau_{lt} \approx 0\\
    \tau_{ll} \approx -4.9 \theta_{ll}\\
    \tau_{lf} \approx 2.6 z_t - 4.3 \theta_t + 3.1 \theta_{rt} + 3.2 \theta_{ll}\\
\end{cases}
\end{equation}
The right thigh's torque aims to balance the torso's angle. Moreover, it has a positive contribution from the z-velocity of the torso.
The right leg's torque, instead, only depends on the torso's angle.
The right foot's torque depends positively on the z-coordinate, the right thigh's angle, and the z-velocity; while depending negatively on the torso's angle and the x-velocity, suggesting that this is the foot taking a step when this expert is called.
The left thigh's torque is $0$, while the left leg's torque only aims at reducing the magnitude of its own angle.
Finally, the left foot shares some similarities with the right foot's torque. However, it does not depend on the velocities, but it does depend on the left leg's angle, suggesting that this joint is moving the foot to prepare it for taking steps in the near future.

\paragraph{Expert 8 ($S_8 \approx -3.1 z_t + 5.4 \theta_t + 7.5 \theta_{rl} - 7.1 \theta_{rf} + 9.7 \theta_{lt} + 8.1 \theta_{ll} + 2.9 \theta_{lf}$)}
This expert is likely to be called when all the angles are positive except for (1) the right thigh's angle (no significant weight), and (2) the right foot, which is likely to be negative.

The policy of this expert shares many similarities with that of Expert 3's, and can be summarized as:
\begin{equation}
\begin{cases}
    \tau_{rt} \approx 4.7 z_t - 11.0 \theta_t - 3.8 \theta_{rt} + 8.3 \theta_{lt} - 8.2 \theta_{ll}\\
    \tau_{rl} \approx 2.1 z_t - 4.7 \theta_t - 7.3 \theta_{ll} \\
    \tau_{rf} \approx -3.1 \theta_t - 3.7 \theta_{rf} + 3.0 \theta_{lt}\\
    \tau_{lt} \approx 2.8 \theta_t - 7.7 \theta_{lt} - 4.3 \theta_{ll}\\
    \tau_{ll} \approx 2.7 z_t - 4.4 \theta_{rl} - 5.7 \theta_{ll}\\
    \tau_{lf} \approx 6.3 z_t - 6.5 \theta_t + 2.4 \theta_{lt} - 2.4 \theta_{ll} + 2.1 v_z\\
\end{cases}
\end{equation}
In fact, we observe that the policy for the right thigh's torque is very similar to that of Expert 3.
Also, the right leg's torque is similar in directions to that of Expert 3, but adds a positive contribution from $z_t$.
A similar reasoning applies to the left thigh's torque, where the additional contribution does not depend on the torso's x-coordinate, but on its angle.
Regarding the other joints, their policies are different.
$\tau_{rf}$ tends to (1) reduce the magnitude of its own angle, and (2) move roughly in the same direction as the difference between the left thigh's and the right thigh's angles.
The left leg's torque tends to move according to the z-coordinate of the torso and to the opposite of the sum of the left leg's and right leg's angles.
Finally, the left foot depends positively on the z-coordinate and negatively on the torso's angle, positively on the left thigh's angle, negatively on the left leg's angle, and positively on the z-velocity.

\subsubsection{Hopper-v4}
\begin{figure*}[ht!]
\centering
\includegraphics[width=1\textwidth]{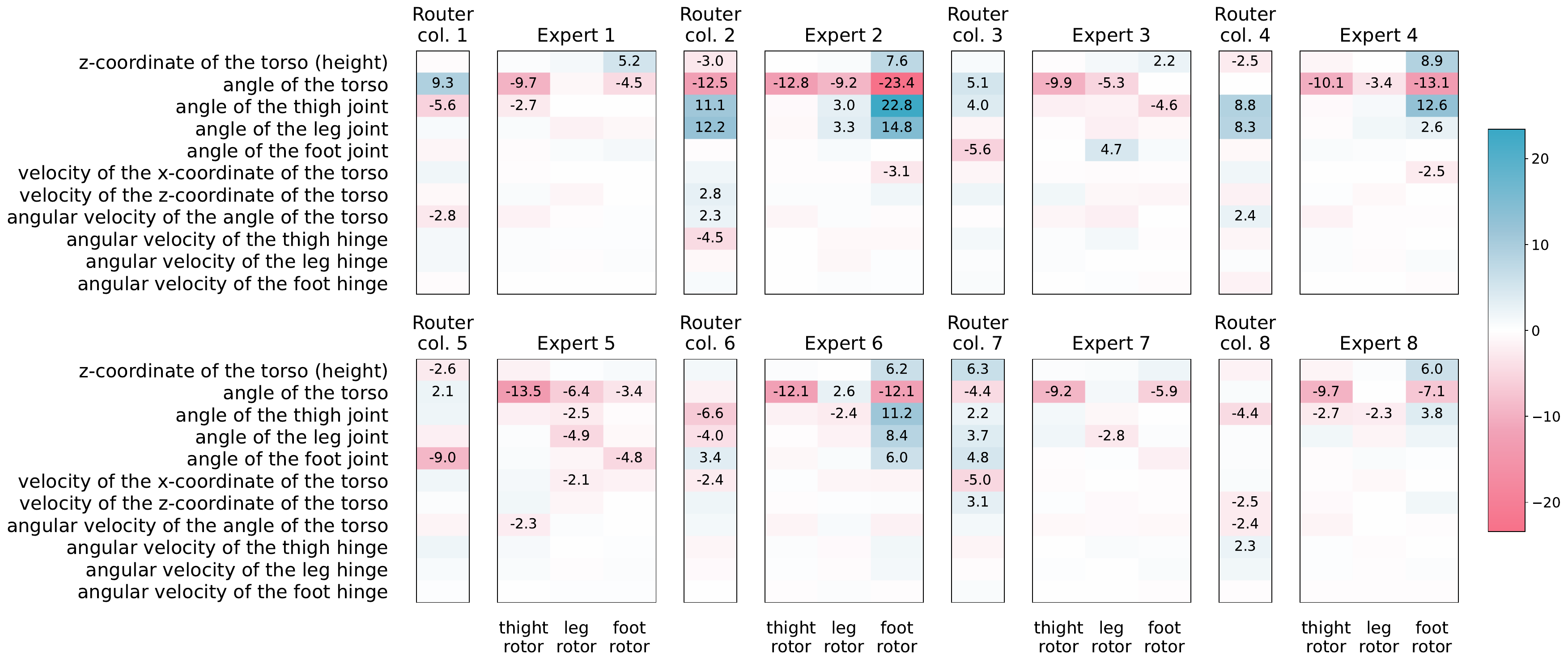}
\caption{\textbf{Hopper-v4}. Visual representation of the learned weights for each expert and of the corresponding column of the router’s weight matrix.}
\label{fig:hopper_weights}
\end{figure*}

In this environment, similarly to Walker2d-v4, our policy uses the following inputs:
\begin{itemize}
    \item z-coordinate of the torso: $z_{t}$
    \item angle of the torso: $\theta_{t}$
    \item angle of thigh: $\theta_{th}$
    \item angle of leg: $\theta_{l}$
    \item angle of foot: $\theta_{f}$
    \item x-velocity of the torso: $v_{x}$
    \item z-velocity of the torso: $v_{z}$
    \item angular velocity of the torso: $\omega_{t}$
    \item angular velocity of right thigh: $\omega_{th}$
    \item angular velocity of right leg: $\omega_{l}$
    \item angular velocity of right foot: $\omega_{f}$
\end{itemize}

to control the following variables:
\begin{itemize}
    \item thigh's torque: $\tau_{th}$
    \item leg's torque: $\tau_{l}$
    \item foot's torque: $\tau_{f}$
\end{itemize}

\paragraph{Expert 1 ($S_1 \approx -9.3 \theta_t - 5.6 \theta_{th} - 2.8 \omega_t$)}
This is the expert that gives the highest weights to $\theta_t$, which suggests that this expert is likely called when $\theta_t$ is positive and $\theta_{th}$ and $\omega_t$ have negative or low-magnitude values. This, for instance, may happen when the hopper is unbalanced and, if not properly controlled, risks falling.

Its policy is:
\begin{equation}
    \begin{cases}
        \tau_{th} \approx -9.7 \theta_t - 2.7 \theta_{th}\\
        \tau_{l} \approx 0\\
        \tau_{f} \approx 5.2 z_t - 4.5 \theta_t\\
    \end{cases}
\end{equation}

The thigh rotor ($\tau_{th}$) aims to reduce the torso's angle by moving the thigh rotor (which is connected to the torso) exploiting a large negative dependency on $\theta_t$. Also, it has an additional contribution from the thigh's angle, which tries to bring the thigh's joint angle towards zero.

The foot rotor, instead, has a positive dependency on $z_t$ which is fairly easy to interpret.
It allows the foot to make jumps by raising the foot's tip during ``landing'', while kicking when the z-coordinate starts increasing (to perform a new jump).
Moreover, it also has a negative dependency on $\theta_t$, which means that its torque has an opposite direction w.r.t. the current thigh angle.

\paragraph{Expert 2 ($S_2 \approx -3.0 z_t - 12.5 \theta_t + 11.1 \theta_{th} + 12.2 \theta_l + 2.8 v_z + 2.3 \omega_t - 4.5 \omega_{th}$)}
The main weights here are those related to $\theta_t, \theta_{th}$, and $\theta_l$. This means that this expert is likely to be queried when the torso's angle is negative and the thigh's and leg's angles are positive.

The policy of this expert can be summarized as:
\begin{equation}
    \begin{cases}
        \tau_{th} \approx -12.8 \theta_t\\
        \tau_{l} \approx -9.2 \theta_t + 3.0 \theta_{th} + 3.3 \theta_l\\
        \tau_{f} \approx 7.6 z_t -23.4 \theta_t + 22.8 \theta_{th} + 14.8 \theta_l - 3.1 v_x\\
    \end{cases}
\end{equation}
This can be interpreted as follows.
The thigh's torque tries to keep the torso balanced (with an angle of $0$ radians).
The leg's torque also has a negative dependency on the torso, while it has two positive contributions from $\theta_{th}$ and $\theta_l$, suggesting that the leg tries to (1) move together with the thigh to balance the robot; and (2) keep moving in the current leg's direction to preserve momentum.
Finally, the foot's torque has a larger number of dependencies.
Besides the positive dependency on $z_t$, which we explained in Expert 1, it has strong dependencies on $\theta_{t}, \theta_{th}$, and $\theta_l$; which suggests that, while this policy tries to make the hopper move the foot in such a way that facilitates ``hops'', its main role is in balancing the robot, together with the other two joints. In fact, it shows the same pattern as $\tau_l$, where we have a negative dependency on $\theta_t$ and positive dependencies on $\theta_{th}$ and $\theta_l$. The last contribution is from $v_x$, which is negative and may mean that another role of this rotor is to reduce the walking speed to avoid falling.

\paragraph{Expert 3 ($S_3 \approx 5.1 \theta_t + 4.0 \theta_{th} - 5.6 \theta_f$)}
This indicates that this expert is likely called when both the torso's angle and the thigh's angle are large, while the foot's angle is negative, which suggests that this expert may have a balancing role.

Its policy can be written as:
\begin{equation}
    \begin{cases}
        \tau_{th} \approx -9.9 \theta_t\\
        \tau_{l} \approx  -5.3 \theta_t + 4.7 \theta_f\\
        \tau_{f} \approx 2.2 z_t -4.6 \theta_{th}\\
    \end{cases}
\end{equation}
We observe that the role of the thigh rotor is to balance the torso, as also happens in Expert 2.
The leg rotor, instead, has a negative contribution from $\theta_t$, while it has a positive contribution from $\theta_f$, which may mean that its role is to serve as a balance between the torso (which affects the thigh rotor) and the foot.
Finally, the foot has the usual dependency on $z_t$, together with a negative dependency on $\theta_{th}$, which may be interpreted as a way to further reduce the angle of the foot to balance the torso and thigh unbalance.

\paragraph{Expert 4 ($S_4 \approx -2.5 z_t + 8.8 \theta_{th} + 8.3 \theta_l + 2.4 \omega_t$)} This suggests that this expert is likely to be called when the z-coordinate of the torso is negative and its angular velocity is positive, while both the thigh's angle and the leg's angle are positive.

The corresponding policy is:
\begin{equation}
    \begin{cases}
        \tau_{th} \approx -10.1 \theta_t\\
        \tau_{l} \approx  -3.4 \theta_t\\
        \tau_{f} \approx 8.9 z_t - 13.1 \theta_t + 12.6 \theta_{th} + 2.6 \theta_l - 2.5 v_x\\
    \end{cases}
\end{equation}
Also in this case, we have a negative dependency on $\theta_t$ for all the control variables.
Moreover, similarly to Expert 2, the foot rotor's torque also depends on $z_t$, the thigh's angle, the leg's angle, and the x-velocity.

\paragraph{Expert 5 ($S_5 \approx - 2.6 z_t + 2.1 \theta_t - 9.0 \theta_f$)}
This rotor is likely to be called when the foot's angle is negative and has a high magnitude, suggesting that this expert's role is to balance the foot's angle.

Its policy can be approximated as:
\begin{equation}
    \begin{cases}
        \tau_{th}\approx -13.5 \theta_t - 2.3 \omega_t\\
        \tau_{l} \approx -6.4 \theta_t - 2.5 \theta_{th} - 4.9 \theta_l - 2.1 v_x\\
        \tau_{f} \approx -3.4 \theta_t - 4.8 \theta_f\\
    \end{cases}
\end{equation}
The torque applied to the thigh's joint is chosen in such a way that it tends to stabilize the torso (as the thigh's joint is connected to it) by dampening the oscillations of both the torso's velocity and the torso's angle.
$\tau_l$, instead, takes into account several variables, namely $\theta_t, \theta_{th}, \theta_l$, and $v_x$. This suggests that this policy tries to balance the robot's structure, not only by trying to compensate its own angle, but also by taking into account most of the other angles in the system, together with an action that is negatively correlated with the x-velocity of the robot ($-2.1 v_x$).
This expert's balancing role is further confirmed by the fact that this is one of the two experts whose $\tau_f$ does not depend on $z_t$.
In fact, here $\tau_f$ only depends on the torso angle and the foot angle.

\paragraph{Expert 6 ($S_6 \approx -6.6 \theta_{th} - 4.0 \theta_l + 3.4 \theta_f - 2.4 v_x$)}
This expert gives high weight to the thigh angle, the leg angle, the foot angle, and the x-velocity.
We can then hypothesize that this expert is queried when the foot has a positive angle, while the thigh and the leg have negative angles, while also having negative x-velocity.

Its policy can be approximated as:
\begin{equation}
    \begin{cases}
        \tau_{th}\approx -12.1 \theta_t\\
        \tau_{l} \approx 2.6 \theta_t - 2.4 \theta_{th}\\
        \tau_{f} \approx 6.2 z_t - 12.1 \theta_t + 11.2 \theta_{th} + 8.4 \theta_l + 6.0 \theta_f\\
    \end{cases}
\end{equation}
Like in most of the experts, also here the thigh's torque is set to be negatively proportional to the torso's angle, to balance it.
The leg's torque, instead, is set to be in accordance with the torso's angle, while balancing the thigh's angle.
Finally, the foot's torque is set similarly (w.r.t. the signs) to Expert 2, with the main difference being the fact that Expert 6 does not aim to reduce x-velocity, but, instead, it aims to preserve the momentum by adding a term proportional to the foot's angle.

\paragraph{Expert 7 ($S_7 \approx 6.3 z_t - 4.4 \theta_t + 2.2 \theta_{th} + 3.7 \theta_l + 4.8 \theta_f - 5.0 v_x + 3.1 v_z$)}
This expert is likely called when $z_t, \theta_{th}, \theta_l, \theta_f, v_z$ are positive, and $\theta_t, v_x$ are negative.

The corresponding policy is:
\begin{equation}
    \begin{cases}
        \tau_{th}\approx -9.2 \theta_t\\
        \tau_{l} \approx -2.8 \theta_l\\
        \tau_{f} \approx -5.9 \theta_t\\
    \end{cases}
\end{equation}
This policy is quite simple, and we can see that $\tau_{th}$, as usual, handles the balancing of the torso; while $\tau_l$ simply stabilizes the leg; and the $\tau_f$ moves the foot in accordance with the thigh, although with a smaller (in magnitude) coefficient.

\paragraph{Expert 8 ($S_8 \approx -4.4 \theta_{th} -2.5 v_z - 2.4 \omega_t + 2.3 \omega_{th}$)}
The scoring function of this expert suggests that it is called when the thigh has a negative angle but a positive angular velocity, while the robot has a negative z-velocity and a negative angular velocity of the torso.

This expert's policy can be summarized as:
\begin{equation}
    \begin{cases}
        \tau_{th}\approx -9.7 \theta_t - 2.7 \theta_{th}\\
        \tau_{l} \approx -2.e \theta_{th}\\
        \tau_{f} \approx 6.0 z_t - 7.1 \theta_t + 3.8 \theta_{th}\\
    \end{cases}
\end{equation}
Here, we observe that $\tau_{th}$ does not depend only on $\theta_t$, but also on $\theta_{th}$, which suggests that this expert tries to stabilize, simultaneously, both the torso and the thigh.
Moreover, $\tau_l$ only depends on $\theta_l$, which means that its main effect is to stabilize the leg.
Finally, $\tau_f$ depends as usual on $z_t$, $\theta_t$, and $\theta_{th}$, suggesting that it also has a stabilization effect on the whole robot.

\subsubsection{Swimmer-v4}
\begin{figure*}[ht!]
\centering
\includegraphics[width=1\textwidth]{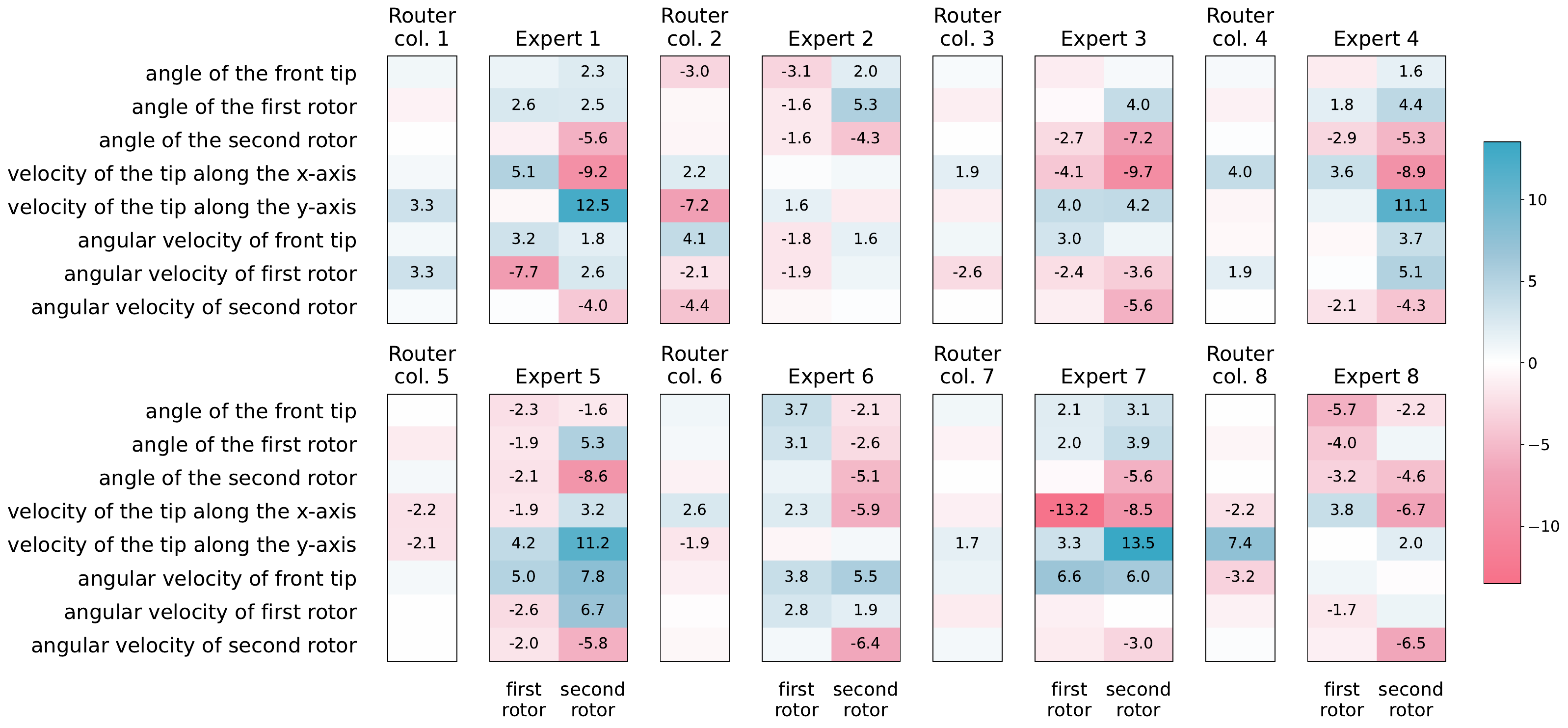}
\caption{\textbf{Swimmer-v4}. Visual representation of the learned weights for each expert and of the corresponding column of the router’s weight matrix.}
\label{fig:swimmer_weights}
\end{figure*}

In Swimmer-v4, the model has access to the following observations:
\begin{itemize}
    \item angle of the front tip: $\theta_f$
    \item angle of the 1st rotor: $\theta_1$
    \item angle of the 2nd rotor: $\theta_2$
    \item x-velocity of the tip: $v_{fx}$
    \item y-velocity of the tip: $v_{fy}$
    \item angular velocity of the tip: $\omega_f$
    \item angular velocity of the 1st rotor: $\omega_1$
    \item angular velocity of the 2nd rotor: $\omega_2$
\end{itemize}
and has to control the following variables:

\begin{itemize}
    \item torque of the first rotor: $\tau_1$
    \item torque of the second rotor: $\tau_2$
\end{itemize}

\paragraph{Expert 1 ($S_1 \, \approx \, 3.3 \, v_{fy} + 3.3 \, \omega_1$)}
Expert 1 is likely to be selected when both the y-coordinate of the tip velocity of Swimmer-v4, and the angular velocity of the first rotor are large, implying a scenario where the swimmer is experiencing both significant sideways movement and active rotation at the first joint. According to its policy:
\begin{equation*}
    \begin{cases}
        \tau_{1} \, \approx \, 2.6 \, \theta_1 + 5.1 \, v_{fx} + 3.2 \, \omega_f -7.7 \, \omega_1, \\
        \tau_{2} \, \approx \,  2.3 \, \theta_f + 2.5 \, \theta_1 - 5.6 \, \theta_2 - 9.2 \, v_{fx} + \\ \qquad \,\, + 12.5 \, v_{fy} + 1.8 \, \omega_{f} + 2.6 \, \omega_1 - 4 \, \omega_2,
    \end{cases}
\end{equation*}
the controller of the first rotor actuated by Expert 1 is primarily influenced by the angle at the first joint, the forward and angular velocity of the tip, and the angular velocity at the first joint itself. The torque $\tau_1$ is positively influenced by the angle at that joint. This suggests that when $\theta_1$ increases, $\tau_1$ increases, to help align the swimmer's segments. The forward velocity of the tip $v_{fx}$ positively contributes to $\tau_1$: as the front of the swimmer moves faster in the x-direction, the torque at the first joint increases. The same can be said about the effect of $\omega_f$: $\tau_1$ grows proportionally to it, possibly to synchronize the motion of the tip and first segment. The only negative dependency is given by $\omega_1$: when the angular rotation of the first rotor is high, the torque decreases, maybe to prevent excessive rotation or instability at the first joint. The control $\tau_2$ on the second rotor is influenced by the angle and velocities of the tip, as well as the angles and angular velocities at both joints. Firstly, we can notice the positive dependency on both $\theta_1$ and $\omega_1$, indicating that movements at the first joint affect the torque at the second joint, creating a coordinated movement. Similar reasoning can be applied for the positive dependency on $\theta_f$ and $\omega_f$. The negative dependency on $\theta_2$ and $\omega_2$ might be interpreted as a way to control and stabilize the motion at this joint. The larger effects are introduced by substantial dependencies on the forward and side tip velocities, negative and positive, respectively. As the tip moves faster forwards, the torque at the second joint decreases. Instead, the faster the tip moves sideways, the more the second joint increases its torque, to adjust the swimmer's orientation and counterbalance the sideways drift. 

\paragraph{Expert 2 ($S_2 \, \approx \, -3 \, \theta_{f} + 2.2 \, v_{fx} - 7.2 \, v_{fy} + 4.1 \, \omega_{f}  - 2.1 \, \omega_{1}  - 4.4 \, \omega_{2}$)} From its score, Expert 2 is likely to be triggered if $\theta_f$ is large and negative, or in cases of decisive forward motion, and not during strong lateral movements in a positive direction. The dependencies of the score on the angular velocities suggest that it is employed when the tip is experiencing quick rotations, while the first and second rotors, conversely, are not in this state. This suggests the expert is likely intended for scenarios where the swimmer is moving forward and the tip is rotating, but the swimmer is relatively stable laterally and not undergoing rapid internal rotations. Its policy is:
\begin{equation*}
    \begin{cases}
        \tau_{1} \, \approx \, -3.1 \, \theta_f - 1.6  \, \theta_1 - 1.6 \, \theta_2  + 1.6 \, v_{fy} + \\ \qquad \quad  - 1.8 \, \omega_{f} -1.9 \, \omega_1, \\
        \tau_{2} \, \approx \, 2 \, \theta_f + 5.3  \, \theta_1 - 4.3 \, \theta_2  + 1.6  \, \omega_{f}.
    \end{cases}
\end{equation*}
Analyzing the first torque $\tau_1$, we can see a $\theta_{f}$ large negative contribution from the tip's angle, likely to avoid excessive forward tilting of the swimmer. Moreover, symmetric negative contributions of $\theta_1$ and $\theta_2$ suggest a stabilizing effect to prevent large deviations in the configuration of the swimmer's body. The policy attempts to react to any lateral drift, through a positive contribution from the sideways velocity. Negative contributions from the angular velocities of the tip and the first joint, likely aim to reduce the torque if the swimmer's tip or first joint is rotating rapidly, promoting stability. Regarding torque $\tau_2$, we see a positive contribution from the tip's angle; a large positive contribution from the angle at the first joint, indicating a strong influence of the first joint on the second, to help coordinate movement between the segments; a smaller positive contribution from the tip's angular velocity, indicating some response to the tip’s rotation but not as dominant as the angle-based terms. Finally, $\tau_2$ presents a negative contribution from the angle at the second joint, possibly helping to avoid excessive bending.

\paragraph{Expert 3 ($S_3 \, \approx \, 1.9 \, v_{fx} - 2.6 \, \omega_1$)} According to the score $S_3$, Expert 3 is selected when Swimmer-v4 is moving fast in the forward direction, while low angular velocity is experienced at the first rotor. The policy associated to Expert 3 is the following:
\begin{equation*}
    \begin{cases}
        \tau_{1} \, \approx \, -2.7 \, \theta_2 - 4.1 \, v_{fx} + 4 \, v_{fy} + 3 \, \omega_{f} -2.4 \, \omega_{1}\\
        \tau_{2} \, \approx \, 4 \, \theta_1 - 7.2 \, \theta_2 - 9.7 \, v_{fx} + 4.2 \, v_{fy}\\ \qquad \quad  - 3.6 \, \omega_1 - 5.6 \, \omega_2 
    \end{cases}
\end{equation*}
In such policy, $\tau_1$ is computed from several components: a component with negative dependency on $\theta_2$ (preventing excessive bending at the second joint); two components with similar dependency on $v_{fx}$ (negative) and $v_{fy}$ (positive), respectively, one counteracting forward movement, and the other adjusting the robot's lateral position; a component linking positively $\tau_1$ and $\omega_f$, acting as a coordinating force between the rotation of the tip and the movement of the first rotor; and a negative contribution from $\omega_1$, possibly to prevent excessive rotation at the first joint. 
Regarding $\tau_2$, instead, we observe a positive dependency on $\theta_1$ and a negative dependency on $\theta_2$, which might be interpreted as contributing to managing the swimmer’s posture. Moreover, negative dependencies on $\omega_1$ and $\omega_2$ reduce the second rotor torque when the first and second joints rotate too quickly, preventing excessive rotation, instability, and uncontrolled movements. Finally, similarly to $\tau_2$ in Expert 1, we find substantial dependencies on the forward and side tip velocities, negative and positive, respectively, following the same interpretation.

\paragraph{Expert 4 ($S_4 \, \approx \, 4 \, v_{fx} + 1.9 \, \omega_1$)} Expert 4 is triggered when Swimmer-v4 is moving forward quickly ($v_{fx}$ us large), or
the first joint is rotating rapidly ($\omega_1$ is large), or both. The policy associated to Expert 4 is:
\begin{equation*}
    \begin{cases}
        \tau_{1} \, \approx \, 1.8 \, \theta_1 - 2.9 \, \theta_2 + 3.6 \, v_{fx} - 2.1 \, \omega_2, \\
        \tau_{2} \, \approx \, 1.6 \, \theta_f + 4.4 \, \theta_1 - 5.3 \, \theta_2 - 8.9 \, v_{fx} + 11.1 \, v_{fy} + \\ \qquad \quad + 3.7 \, \omega_f - 5.1 \, \omega_1 - 4.3 \, \omega_2.
    \end{cases}
\end{equation*}
In both $\tau_1$ and $\tau_2$, we can notice a positive dependency on $\theta_1$, contributing to the robot's posture, and a negative dependency on $\theta_2$, that contributes to avoiding excessive bending. Both effects are stronger in $\tau_2$. Moreover, in $\tau_1$ the positive dependency on the forward velocity of the tip $v_{fx}$ might help manage or stabilize the forward propulsion, while the negative one from $\omega_2$ potentially prevents excessive rotation of the body. In $\tau_2$, instead, the positive relations with $\theta_f$ and $\omega_f$ can show an attempt to coordinate the behavior at the second rotor with the behavior of the tip. The negative dependency on $\omega_1$ and $\omega_2$ suggests once more a counteraction against excessive rotation of the body. Once more, similarly to $\tau_2$ in Expert 1 and Expert 3, we find substantial dependencies on the forward and side tip velocities, negative and positive, respectively, following the same interpretation. 

\paragraph{Expert 5 ($S_5 \, \approx \, - 2.2 \, v_{fx} - 2.1 \, v_{fy}$)} According to its score, Expert 5 is selected when this policy is triggered when the swimmer is moving in a backward and/or leftward direction. According to its policy:
\begin{equation*}
    \begin{cases}
        \tau_{1} \, \approx \, -2.3 \, \theta_f - 1.9 \, \theta_1 - 2.1 \, \theta_2 - 1.9 \, v_{fx} + 4.2 \, v_{fy} + \\ \qquad \quad + 5 \, \omega_f - 2.6 \, \omega_1 - 2 \, \omega_2, \\
        \tau_{2} \, \approx \, -1.6 \, \theta_f + 5.3 \, \theta_1 - 8.6 \, \theta_2 + 3.2 \, v_{fx} + 11.2 \, v_{fy} + \\ \qquad \quad + 7.8 \, \omega_f + 6.7 \, \omega_1 - 5.8 \, \omega_2, 
    \end{cases}
\end{equation*}
the torque $\tau_{1}$ is affected by a negative dependency on the tip's angle $\theta_f$, the first joint's angle $\theta_1$, and the second joint's angle $\theta_2$, likely to prevent excessive bending or undesired rotations in the three points. Regarding the angular velocities, $\tau_{1}$ is directly proportional to $\omega_f$, while inversely proportional to $\omega_1$ and $\omega_2$, ensuring coordination between the first rotor motion and the rotation of the tip, while counterbalancing excessive rotation at the two controlled joints. The negative dependency on $v_{fx}$ could be interpreted as an attempt to reduce or invert propulsion, given that this controller, as previously mentioned, is likely employed when the swimmer is moving backward. The direct dependency on $v_{fy}$, helps to lightly correct the lateral movement.
The torque $\tau_2$ at the second joint, similarly to $\tau_1$, shows a negative dependency on the tip's angle $\theta_f$. The positive contribution from $\theta_1$ and the larger negative contribution from  $\theta_2$ suggest that the first joint’s angle increases $\tau_2$ while the second joint’s angle decreases it, balancing the swimmer’s posture. We can notice as well that $\tau_2$, although directly proportional to $v_{fx}$, and hence amplifying its motion in the current direction, has a stronger coefficient than $\tau_2$ reacting to $v_{fy}$, useful for more effective counterbalancing of the lateral movement suggested by $S_5$. The contributions to $\tau_2$ affected by the angular velocities are such that $\tau_2$ is directly proportional to $\omega_f$ and $\omega_1$, coordinating the movement of the second joint with the tip and the first joint, while controlling excessive rotation through a negative dependency on $\omega_2$.

\paragraph{Expert 6 ($S_6 \, \approx \, 2.6 \, v_{fx} - 1.9 \, v_{fy}$)} This expert is called by the router when Swimmer-v4, while is moving forward quickly and/or when it is drifting leftward. Its policy is:
\begin{equation*}
    \begin{cases}
        \tau_{1} \, \approx \, 3.7 \, \theta_f + 3.1 \, \theta_1 + 2.3 \, v_{fx} + \\ \qquad \quad + 3.8 \, \omega_f + 2.8 \, \omega_1 \\
        \tau_{2} \, \approx \, -2.1 \, \theta_f - 2.6 \, \theta_1 - 5.1 \, \theta_2 - 5.9 \, v_{fx} + \\ \qquad \quad + 5.5 \, \omega_f + 1.9 \, \omega_1 - 6.4 \, \omega_2 
    \end{cases}
\end{equation*}
The positive dependencies of $\tau_1$ on $\theta_f$ and $\theta_1$ attempt to ensure that the robot is in the correct position and aligned with the desired direction. The positive dependencies on $v_{fx}$, $\omega_f$, and $\omega_1$, likely try to reinforce the forward movement of Swimmer-v4, coordinate the tip's movement, and stabilize/enhance its motion, respectively. 
Analyzing $\tau_2$, we can notice the negative dependency on $\theta_f$,  $\theta_1$, and $\theta_2$, which we can interpret as reducing excessive motion at the tip, and bending or unwanted joint rotations. Moreover, the negative contribution from $v_{fx}$ means that when the forward velocity grows, it decreases $\tau_2$, which slows down the swimmer, reducing excessive forward propulsion. Finally, the positive dependency of $\tau_2$ on $\omega_f$ and $\omega_1$, and the negative one on $\omega_2$ can be interpreted equally to the ones spotted in $\tau_2$, when controlled by Expert 5.

\paragraph{Expert 7 ($S_7 \, \approx \, 1.7 \, v_{fy}$)} Score $S_7$ will be large when $v_{fy}$
  (the sideways velocity of the swimmer's tip) is large and positive, meaning that Expert 7 is selected when Swimmer-v4 is moving significantly towards the right. The associated policy is:
\begin{equation*}
    \begin{cases}
        \tau_{1} \, \approx \, 2.1 \, \theta_f + 2 \, \theta_1 - 13.2 \, v_{fx} + 3.3 \, v_{fy} + 6.6 \, \omega_f, \\
        \tau_{2} \, \approx \, 3.1 \, \theta_f + 3.9 \, \theta_1 -5.6 \, \theta_2 - 8.5 \, v_{fx} + 13.5 \, v_{fy} + \\ \qquad \quad + 6 \,\omega_f - 3 \, \omega_2.
    \end{cases}
\end{equation*}
According to this policy, the torque $\tau_1$ increases with the angles at the tip $\theta_f$ and first joint $\theta_1$, as well as with the sideways velocity $v_{fy}$ and tip’s angular velocity $\omega_f$. This helps to correct the swimmer's posture and manage its rightward drift, while the negative dependency on $v_{fx}$ might be counteracting excessive forward velocity. The second torque $\tau_2$ increases with similar parameters ($\theta_f$, $\theta_1$, $v_{fy}$, and $\omega_f$) but decreases with forward velocity $v_{fx}$, second joint’s angle $\theta_2$, and angular velocity at the second joint $\omega_2$. This helps maintain stability, control lateral movement, and prevent excessive bending or rotational instability.

\paragraph{Expert 8 ($S_8 \, \approx \, - 2.2 \, v_{fx} + 7.4 \, v_{fy} -3.2 \, \omega_{f}$)} According to $S_8$, Expert 8 is selected when Swimmer-v4 has a high sideways velocity $v_{fy}$ (and hence significant sideways movement, particularly to the right) and possibly negative forward velocity $v_{fx}$, and tip angular velocity $\omega_f$. Its policy is:
\begin{equation*}
    \begin{cases}
        \tau_{1} \, \approx \, -5.7 \, \theta_f - 4 \, \theta_1 -3.2 \, \theta_2 + 3.8 \, v_{fx}  \, -1.7 \, \omega_1, \\
        \tau_{2} \, \approx \, -2.2 \, \theta_f - 4.6 \, \theta_2 - 6.7 \, v_{fx} + 2 \, v_{fy} - 6.5 \, \omega_2.
    \end{cases}
\end{equation*}
From the policy, we observe that the first rotor's torque $\tau_1$ primarily decreases with larger angles at the tip and joints ($\theta_f$, $\theta_1$, and $\theta_2$), as well as with the angular velocity at the first joint $\omega_1$. It increases with forward velocity $v_{fx}$. This behavior could help maintain forward motion, while counteracting excessive lateral and rotational movement. The second joint's torque $\tau_2$ also decreases with larger angles $\theta_f$, and $\theta_2$, and forward velocity $v_{fx}$, and significantly with the angular velocity at the second joint $\omega_2$. It increases with the sideways velocity $v_{fy}$, which may be to correct or manage lateral drift.

\subsubsection{HalfCheetah-v4}

The model has access to the following pieces of information:
\begin{itemize}
    \item z-coordinate of the torso: $z_{t}$
    \item angle of the torso: $\theta_{t}$
    \item angle of the back thigh: $\theta_{bt}$
    \item angle of the back shin: $\theta_{bs}$
    \item angle of the back foot: $\theta_{bf}$
    \item angle of the front thigh: $\theta_{ft}$
    \item angle of the front shin: $\theta_{fs}$
    \item angle of the front foot: $\theta_{ff}$
    \item x-velocity of the torso: $v_{xt}$
    \item z-velocity of the torso: $v_{zt}$
    \item angular velocity of the torso: $\omega_{t}$
    \item angular velocity of the back thigh: $\omega_{bt}$
    \item angular velocity of the back shin: $\omega_{bs}$
    \item angular velocity of the back foot: $\omega_{bf}$
    \item angular velocity of the front thigh: $\omega_{ft}$
    \item angular velocity of the front shin: $\omega_{fs}$
    \item angular velocity of the front foot: $\omega_{ff}$
\end{itemize}

\begin{itemize}
    \item Torque applied on the back thigh: $\tau_{bt}$
    \item Torque applied on the back shin: $\tau_{bs}$
    \item Torque applied on the back foot: $\tau_{bf}$
    \item Torque applied on the front thigh: $\tau_{ft}$
    \item Torque applied on the front shin: $\tau_{fs}$
    \item Torque applied on the front foot: $\tau_{ff}$
\end{itemize}

\paragraph{Expert 1 ($S_1 \approx -4.6 z_t + 2.3 \theta_t - 2.4 \theta_{bs}$)}

Its policy is:
\begin{equation}
    \begin{cases}
        \tau_{bt} \approx 0 \\
        \tau_{bs} \approx -3.1 z_t \\
        \tau_{bf} \approx -3.0 z_t \\
        \tau_{ft} \approx -9.6 z_t \\
        \tau_{fs} \approx 7.0 z_t \\
        \tau_{ff} \approx 8.2 z_t
    \end{cases}
\end{equation}

It is interesting to note that this expert only relies on the z-coordinate of the front tip.
It does not apply any torque to the back thigh rotor, while, for the other two back rotors ($\tau_{bs}, \tau_{bf}$) and the front thigh rotor ($\tau_{ft}$), it applies a torque that has a negative dependency on the z-coordinate. For the remaining two ($\tau_{fs}, \tau{ff}$), it applies a torque that is proportional to the z-coordinate.

\paragraph{Expert 2 ($S_2 \approx 8.6 z_t - 4.6 \theta_t - 3.1 \theta_{bt} + 6.9 \theta_{ft} + 4.4 \theta_{fs}$)} This suggests that this expert is likely to be queried: (1) when the z-coordinate of the tip, the angle of the front thigh rotor, and the angle of the front shin rotor are large; and (2) when the z-coordinate angle of the tip, together with the angle of the back thigh rotor, are negative.

Its policy can be approximated as:
\begin{equation}
    \begin{cases}
        \tau_{bt} \approx -23.1 z_t - 3.6 \theta_t - 2.1 \theta_{ft}\\
        \tau_{bs} \approx -4.1 z_t\\
        \tau_{bf} \approx -3.4 \theta_t\\
        \tau_{ft} \approx -14.4 z_t + 7.1 \theta_t\\
        \tau_{fs} \approx 8.6 z_t\\
        \tau_{ff} \approx 3.6 z_t - 2.5 \theta_t - 2.4 \theta_{ft}\\
    \end{cases}
\end{equation}

Also in this case, most of the outputs have a dependency on the z-coordinate of the tip, which seems to contain a significant amount of information for solving this task.

We observe that $\tau_{bt}, \tau_{bs}$, and $\tau_{ft}$ have a negative dependency on $z_t$. On the other hand, $\tau_{fs}$ and $\tau_{ff}$ have a positive dependency on $z_t$.
Moreover, we also observe that the z-coordinate angle of the tip has an important contribution to this expert's policy.
In fact, $\tau_{bt}, \tau{bf}$ and $\tau_{ff}$ have a negative dependency on $\theta_t$, while $\tau_{ft}$ has a positive dependency on it.

\paragraph{Expert 3 ($S_3 \approx -8.1 z_t + 2.4 \theta_{bt}$)}
This suggests that this expert, similarly to Expert 1, is called when $z_t$ has a negative value. However, in this case $S_3$ does not depend on $\theta_t$ and $\theta_{bs}$, but depends on $\theta_{bt}$.

Its policy is:
\begin{equation}
    \begin{cases}
        \tau_{bt} \approx -2.9 z_t\\
        \tau_{bs} \approx -6.0 z_t\\
        \tau_{bf} \approx 0 \\
        \tau_{ft} \approx -25.3 z_t + 7.8 \theta_t\\
        \tau_{fs} \approx 11.7 z_t\\
        \tau_{ff} \approx 23.7 z_t - 4.9 \theta_t\\
    \end{cases}
\end{equation}

This expert shares many similarities with Expert 1: $\tau_{bs}$ and $\tau_{ft}$ negatively depend on $z_t$, while $\tau_{fs}$ and $\tau{ff}$ positively depend on $z_t$.
However, there are differences: $\tau_{bt}$ is not zero, but instead it has a negative dependency on $z_t$, while here $\tau_{bf}$ is zero.
Moreover, in this expert, we have additional dependencies on $\theta_t$ in $\tau_{ft}$ and $\tau{ff}$.

\paragraph{Expert 5 ($S_5 \approx 17.9 z_t - 15.6 \theta_t - 3.4 \theta_{ft} - 3.4 \theta_{fs} + 4.0 v_{zt}$)} While this is more complex than most of the other experts, it is worth noticing that it heavily depends on $z_t$ and $\theta_t$, while the other terms have much smaller weights.

The corresponding policy is:
\begin{equation}
    \begin{cases}
        \tau_{bt} \approx -4.5 z_t\\
        \tau_{bs} \approx 6.2 z_t\\
        \tau_{bf} \approx -4.7 z_t\\
        \tau_{ft} \approx 5.2 z_t - 2.7 \theta_t\\
        \tau_{fs} \approx 7.5 z_t - 5.9 \theta_t\\
        \tau_{ff} \approx 4.2 z_t - 2.5 \theta_t\\
    \end{cases}
\end{equation}

We observe that all the joints depend on the z-coordinate of the tip (either positively or negatively).
Moreover, all the front joints have similar behaviors: positively depending on $z_t$, while negatively depending on $\theta_t$.

\paragraph{Expert 6 ($S_6 \approx 5.5 z_t - 7.6 \theta_t - 5.0 \theta_{ft} - 2.7 \theta_{fs}$)}
Here the magnitude of the terms is more uniformly distributed. We observe that the magnitude of the weights assigned to $z_t$ and $\theta_t$ is comparable with those of $\theta_{ft}$ and $\theta_{fs}$, which suggests that this expert is preferred over Expert 5 when $\theta_{ft}$ and $\theta_{fs}$ have negative values and they are larger (in magnitude) than $z_t$ and $\theta_t$.

\begin{equation}
    \begin{cases}
        \tau_{bt} \approx -15.0 z_t + 2.3 \theta_t\\
        \tau_{bs} \approx 0\\
        \tau_{bf} \approx -3.2 z_t - 4.6 \theta_t\\
        \tau_{ft} \approx -12.7 z_t \\
        \tau_{fs} \approx 12.2 z_t - 5.2 \theta_t\\
        \tau_{ff} \approx 15.5 z_t - 7.4 \theta_t - 2.5 \theta_{ft}\\
    \end{cases}
\end{equation}

In this case we also observe a negative dependency on $z_t$ on the back rotors (except $\tau_{bs}$) and $\tau_{ft}$.
However, this is the only expert that has a positive dependency on $\theta_t$ in $\tau_{bt}$, even though its weights are an order of magnitude smaller than the dependency on $z_t$.
All the other rotors, except for $\tau_{bs}$ and $\tau_{ft}$ have instead a negative dependency on $\theta_t$.
Finally, it is worth pointing out that $\tau_{ff}$ has a minor dependency on $\theta_{ft}$.

\paragraph{Expert 8 ($S_8 \approx -2.2 z_t - 7.9 \theta_t - 3.6 \theta{bt} + 3.2 \theta_{bs} + 4.1 \theta_{ft}$)}

This expert policy can be summarized as:
\begin{equation}
    \begin{cases}
        \tau_{bt} \approx -9.9 z_t - 4.6 \theta_t\\
        \tau_{bs} \approx -5.6 z_t - 6.9 \theta_t - 3.6 \theta_{bs}\\
        \tau_{bf} \approx -9.9 z_t + 5.4 \theta_t\\
        \tau_{ft} \approx 0\\
        \tau_{fs} \approx 7.7 z_t - 5.1 \theta_t\\
        \tau_{ff} \approx 2.9 z_t - 3.2 \theta_t\\
    \end{cases}
\end{equation}

Interestingly, also in this case we have a positive dependency on $z_t$ in the last two joints ($\tau_{fs}, \tau_{ff}$), which is a common feature of all the experts.
We also note that most of the output variables negatively depend on $\theta_t$, except for $\tau{bf}$ and $\tau_{ft}$.
Finally, it is worth noting that $\tau_{bs}$ also has a negative dependency on $\theta_{bs}$, which is the angle of the same joint the torque is applied to. This may indicate that there is a term that tries to dampen the oscillations of the back shin joint.

\begin{figure*}[ht!]
\centering
\includegraphics[width=1\textwidth]{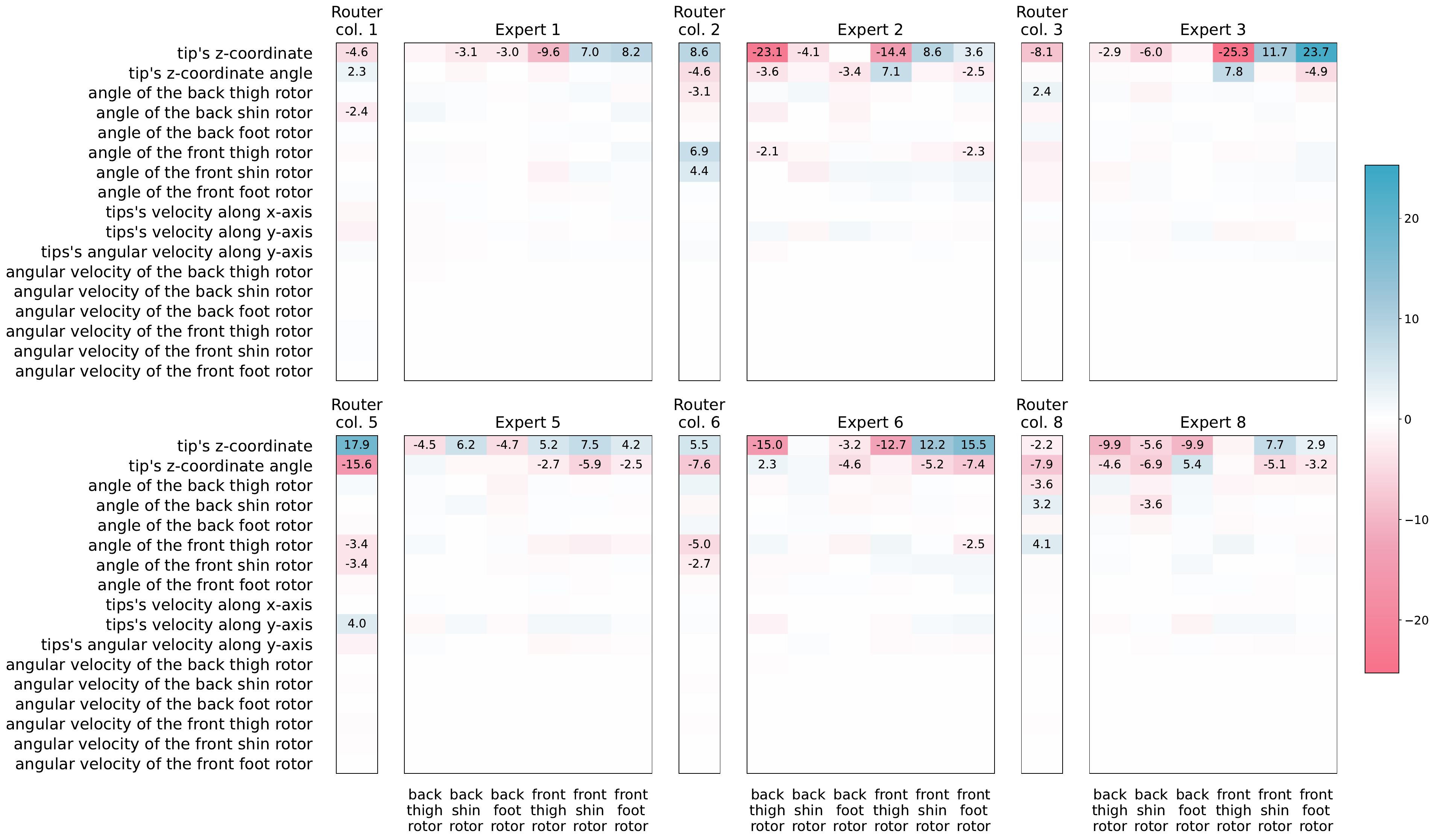}
\caption{\textbf{HalfCheetah-v4}. Visual representation of the learned weights for each expert and of the corresponding column of the router’s weight matrix.}
\label{fig:halfcheetah_weights}
\end{figure*}

\subsubsection{Ant-v4}

The model can access the following inputs:
\begin{itemize}
    \item z-coordinate of the torso (center): $z_{t}$
    \item x-orientation of the torso (center): $\psi_{tx}$
    \item y-orientation of the torso (center): $\psi_{ty}$
    \item z-orientation of the torso (center): $\psi_{tz}$
    \item real term of the quaternion: $\omega_{t}$
    \item angle between torso and the first link on front left: $\theta_{fl}$
    \item angle between the two links on the front left: $\theta_{fl2}$
    \item angle between torso and the first link on the front right: $\theta_{fr}$
    \item angle between the two links on the front right: $\theta_{fr2}$
    \item angle between torso and the first link on the back left: $\theta_{bl}$
    \item angle between the two links on the back left: $\theta_{bl2}$
    \item angle between torso and the first link on back right: $\theta_{br}$
    \item angle between the two links on the back right: $\theta_{br2}$
    \item x-coordinate velocity of the torso: $v_{x}$
    \item y-coordinate velocity of the torso: $v_{y}$
    \item z-coordinate velocity of the torso: $v_{z}$
    \item x-coordinate angular velocity of the torso: $\omega_{x}$
    \item y-coordinate angular velocity of the torso: $\omega_{y}$
    \item z-coordinate angular velocity of the torso: $\omega_{z}$
    \item angular velocity of the angle between torso and front left link: $\omega_{fl}$
    \item angular velocity of the angle between front left links: $\omega_{fl2}$
    \item angular velocity of the angle between torso and front right link: $\omega_{fr}$
    \item angular velocity of the angle between front right links: $\omega_{fr2}$
    \item angular velocity of the angle between torso and back left link: $\omega_{bl}$
    \item angular velocity of the angle between back left links: $\omega_{bl2}$
    \item angular velocity of the angle between torso and back right link: $\omega_{br}$
    \item angular velocity of the angle between back right links: $\omega_{br2}$
\end{itemize}
and has to control the following variables:
\begin{itemize}
    \item Torque applied on the rotor between the torso and back right hip: $\tau_{br}$
    \item Torque applied on the rotor between the back right two links: $\tau_{br2}$
    \item Torque applied on the rotor between the torso and front left hip: $\tau_{fl}$
    \item Torque applied on the rotor between the front left two links: $\tau_{fl2}$
    \item Torque applied on the rotor between the torso and front right hip: $\tau_{fr}$
    \item Torque applied on the rotor between the front right two links: $\tau_{fr2}$
    \item Torque applied on the rotor between the torso and back left hip: $\tau_{bl}$
    \item Torque applied on the rotor between the back left two links: $\tau_{bl2}$
\end{itemize}

\paragraph{Expert 1 ($-2.3 \omega_t + 1.9 \theta_{bl}$)}
This expert is likely queried when the scalar part of the quaternion ($\omega_t$) is negative, and when the angle between the torso and the first link on the back left is positive.

The policy can then be summarized as:
\begin{equation}
    \begin{cases}
        \tau_{br} \approx 1.9 \psi_{ty}\\
        \tau_{br2} \approx 2.5 z_t - 1.7 \psi_{tx} + 2.4 \psi_{ty} + 2.3 \psi_{tz} - 4.0 \omega_t\\
        \tau_{fl} \approx - 2.4 z_t - 1.6 \theta_{fl}\\
        \tau_{fl2} \approx -1.6 z_t - 3.0 \psi_{ty} + 3.9 \psi_{tz} - 2.0 \theta_{fl2} \\
        \tau_{fr} \approx -1.7 z_t + 3.2 \psi_{tz} + 2.3 \omega_t \\
        \tau_{fr2} \approx -1.7 \theta_{fr2}\\
        \tau_{bl} \approx 10.7 z_t - 2.7 \psi_{tx} + 3.7 \psi_{ty} - 5.7 \psi_{tz} - 2.7 \theta_{bl}\\
        \tau_{bl2} \approx 2.1 \psi_{ty}\\
    \end{cases}
\end{equation}

We observe that this expert often exploits the z-coordinate of the torso, which is used in 5 variables over 8, often with a significant magnitude.
Moreover, we also observe the dependencies of some variables w.r.t. the quaternion of the torso ($\omega_t, \psi_{tx}, \psi_{ty}, \psi_{tz}$.
In fact, $\tau_{br2}$ has a negative dependency on the x-orientation and $\omega_t$ of the torso, while it has a positive dependency on the other two orientations; $\tau_{fl2}$ has a negative dependency on $\psi_{ty}$ while having a positive dependency on $\psi_{tz}$; $\tau_{fr}$ depends positively on both $\psi_{tz}$ and $\omega_t$; $\tau_{bl}$ depends negatively on $\psi_{tx}$ and $\psi_{tz}$, while having a positive contribution from $\psi_{ty}$; finally, $\tau_{bl2}$ is directly proportional to $\psi_{ty}$.

\paragraph{Expert 2 ($S_2 \approx 2.7 z_t - 3.2 \psi_{tz} - 1.7 \omega_t + 2.0 v_z$)}
This expert is likely to be called when both the z-coordinate and the z-velocity are high, while there are also negative rotations around the z-axis.

Its policy is:
\begin{equation}
    \begin{cases}
        \tau_{br} \approx -2.9 \psi_{ty} + 1.8 \psi_{tz}\\
        \tau_{br2} \approx -2.1 \psi_{tx} + 2.6 \psi_{ty} - 2.8 \omega_t - 1.9 \theta_{br2}\\
        \tau_{fl} \approx 2.5\psi_{ty}\\
        \tau_{fl2} \approx -1.9 \psi_{tz}\\
        \tau_{fr} \approx 3.1 \psi_{tz} - 1.6 \theta_{fr} \\
        \tau_{fr2} \approx 2.3 \psi_{tz}\\
        \tau_{bl} \approx -2.0 \psi_{ty} + 2.0 \psi_{tz}\\
        \tau_{bl2} \approx 2.0 \psi_{tx} + 1.6 \omega_t\\
    \end{cases}
\end{equation}

It can be interpreted as follows.
$\tau_{br}$ is composed of two contributions: the first one makes it move opposed to the angle of the center of the torso w.r.t. the y-axis; while the second one makes it move according to the angle of the torso's center w.r.t. the z-axis.
$\tau_{br2}$ has negative dependencies on $\psi_{tx}$ and $\omega_t$, while having a positive contribution from $\psi_{ty}$; finally, it depends (negatively) on the angle of its own joint, which suggests that one of the roles of this policy is to ``stabilize'' $\theta_{br2}$.
$\tau_{fl}$ only depends on the rotation of the torso w.r.t. the y axis.
$\tau_{fl2}$, on the other hand, only has a negative contribution from the z-orientation of the torso, which suggests that the last two joints may be preparing for taking a step in the following timesteps.
$\tau_{fr}$ has two main contributions: one from the z-orientation of the torso, and one from the angle of the same joint it is acting upon (of negative phase).
$\tau_{fr2}$, instead, only depends on the z-orientation of the torso.
These two last torques together seem to move the right leg to make a step (thus reducing $\psi_{tz}$ in the following timesteps).
$\tau_{bl}$ has a similar behavior to $\tau_{br}$.
Finally, $\tau_{bl2}$ has two positive contributions: one from $\psi_{tx}$, and one from $\omega_t$.

\paragraph{Expert 3 ($S_3 \approx = 3.5 \psi_{ty} - 2.7 \omega_t - 2.7 \theta_{bl}$)}
This expert is likely called when there are large rotations of the torso around the y-axis, and, simultaneously, the angle between the torso and the first link on the back left of the robot is negative.

The policy of this expert is:
\begin{equation}
    \begin{cases}
        \tau_{br} \approx 1.7 \psi_{tz} + 1.6 \omega_t\\
        \tau_{br2} \approx -1.9 z_t + 1.6 \psi_{tx} + 2.7 \psi_{tz}\\
        \tau_{fl} \approx -2.4 \psi_{tx} + 8.1 \psi_{ty} + 2.5 \psi_{tz} - 2.5 \omega_t - 1.6 \theta_{fl} - 2.6 \theta_{bl}\\
        \tau_{fl2} \approx -1.8 \psi{tz}\\
        \tau_{fr} \approx 2.9 \psi_{ty} + 3.2 \theta_{tz}\\
        \tau_{fr2} \approx 1.6 \psi_{ty}\\
        \tau_{bl} \approx 2.0 \psi_{tx} + 4.8 \psi_{tz} + 2.4 \omega_t\\
        \tau_{bl2} \approx -2.5 z_t + 2.8 \psi_{tx} - 2.9 \psi_{ty} + 2.1 \psi_{tz}\\
    \end{cases}
\end{equation}
The first control variable, $\tau_{br}$, depends mainly on the z-orientation of the torso (including contributions due to the real part of the quaternion, $\omega_t$). The controller of the back-right leg ($\tau_{br2}$), instead, has a negative dependency on $z_t$, and a positive one from the x-orientation of the torso.
$\tau_{fl}$ has a more complex output, which depends on the whole orientation of the torso (i.e., the full quaternion ($\omega_t, \psi_{tx}, \psi_{ty}, \psi_{tz}$)) with different coefficients; plus a negative contribution depending on $\theta_{fl}$ and another on $\theta_{bl}$.
$\tau_{fl2}$, similarly to $\tau_{br}$ only depends on the z-orientation of the torso. $\tau{fr}$, instead, depends on both the y-orientation and the z-orientation of the torso.
$\tau_{fr2}$ is directly proportional to the y-orientation of the torso.
$\tau_{bl}$ depends on the x-orientation and the y-orientation of the torso (including $\omega_t$), and, finally, $\tau_{bl2}$ depends on all the three orientations plus the z-coordinate of the torso.

\paragraph{Expert 4 ($S_4 \approx -2.0 z_t - 1.9 \psi_{ty} - 3.4 \psi_{tz} - 1.6 \theta_{br2}$)}
The scoring function of this expert suggests that it is mainly queried when (1) the z-coordinate of the torso is negative, (2) the y-orientation and the z-orientation of the torso are both negative, and (3) the angle between the two links of the right leg is negative.

Its policy can be summarized as:
\begin{equation}
    \begin{cases}
        \tau_{br} \approx 0\\
        \tau_{br2} \approx -1.8 \psi_{tx} + 3.6 \psi_{ty} - 1.6 \psi_{tz} - 2.7 \omega_t\\
        \tau_{fl} \approx -3.5 \psi_{tz} - 1.9 \theta_{fl} - 1.7 \theta_{br} + 1.7 v_z\\
        \tau_{fl2} \approx -2.0 \psi_{ty} - 2.8 \psi_{tz}\\
        \tau_{fr} \approx 0\\
        \tau_{fr2} \approx 2.1 \psi_{tz}\\
        \tau_{bl} \approx 2.4 \psi_{ty}\\
        \tau_{bl2} \approx 1.6 \psi_{tx} + 2.4 \psi_{ty} + 2.8 \omega_t\\
    \end{cases}
\end{equation}
This expert, interestingly, does not apply any significant torque to two joints: $\tau_{br} and \tau_{fr}$.
$\tau_{br2}$, on the other hand, depends on all the terms of the quaternion.
$\tau_{fl}$ depends on: the z-orientation of the torso, the angle between the torso and first link on the front left, the angle between the torso and the first link on the back right, and the z-velocity.
$\tau_{fl2}$ depends on the y-orientation and the z-orientation of the torso.
$\tau_{fr2}$ only depends on the z-orientation.
Similarly, $\tau_{bl}$ depends on the y-orientation, while $\tau_{bl2}$ depends on both the x-orientation and the z-orientation.

\paragraph{Expert 6 ($S_6 \approx -5.5 z_t - 3.6 \psi_{tx} - 5.5 \psi_{ty} + 9.7 \psi_{tz} -1.6 \omega_t - 2.4 \theta_{fl} + 2.8 \theta_{fr} - 2.8 v_z$)}
This expert may be called when the z-coordinate of the torso is negative, both the x-orientation and the y-orientation are negative, while the z-orientation is positive, the angle between the torso and the first link is positive, and the z-velocity is negative.

The policy of this expert is:
\begin{equation}
    \begin{cases}
        \tau_{br} \approx 1.7 \psi_{ty}\\
        \tau_{br2} \approx -1.7 \omega_t\\
        \tau_{fl} \approx -4.7 z_t + 4.3 \psi_{tz} + 3.7 \omega_t\\
        \tau_{fl2} \approx 1.8 \psi_{ty} - 1.6 \psi_{tz} - 2.1 \theta_{fl}\\
        \tau_{fr} \approx 1.9 \psi_{tz}\\
        \tau_{fr2} \approx 0\\
        \tau_{bl} \approx 3.9 z_t - 2.6 \psi_{tx}\\
        \tau_{bl2} \approx -2.1 z_t - 1.9 \psi_{tz}\\
    \end{cases}
\end{equation}
Similarly to Expert 5, most of the control variables depend on one or two orientations from the torso's quaternion.
Moreover, we observe that $\tau_{fl}$, $\tau_{bl}$, and $\tau_{bl2}$ depend on the z-coordinate of the torso, while $\tau_{fl2}$ has a (negative) dependency on the same angle it applies its torque to, which suggests that it tries to move it close to an angle of 0.

\paragraph{Expert 7 ($S_7 \approx 3.1 z_t - 2.3 \omega_t - 1.7 \theta_{bl}$)}
This expert is likely to be queried when the z-coordinate is large, or when the torso's orientation has a negative $\omega_t$, while simultaneously having a negative angle between the torso and the first back-left link.

Its policy is:
\begin{equation}
    \begin{cases}
        \tau_{br} \approx -4.3 \psi_{ty}\\
        \tau_{br2} \approx -1.9 z_t + 3.5 \psi_{ty} + 6.0 \psi_{tz} \\
        \tau_{fl} \approx 1.7 z_t - 1.7 \psi_{ty} + 2.1 \psi_{tz} - 2.3 \theta_{fl} - 1.8 \theta_{bl}\\
        \tau_{fl2} \approx -2.8 \psi_{ty} + 2.0 \psi_{tz}\\
        \tau_{fr} \approx 0\\
        \tau_{fr2} \approx 1.9 \psi_{tx}\\
        \tau_{bl} \approx 4.1 \psi_{ty} - 2.0 \psi_{tz}\\
        \tau_{bl2} \approx -1.8 z_t + 2.2 \psi_{tx} + 2.3 \psi_{ty} - 1.6 \theta_{bl2}\\
    \end{cases}
\end{equation}
We observe that most of the dependencies are on orientation components, which suggests a high importance for these features also for this expert.
Moreover, we observe dependencies between $\tau_{fl}$ and $\theta_{fl}$ (which suggests that it tries to bring this angle to 0), $\tau_{fl}$ and $\theta_{bl}$; and $\tau_{bl2}$ and $\theta_{bl2}$ (which, similarly to $\tau_{fl}$, suggests that it tends to move this angle towards 0).

\paragraph{Expert 8 ($S_8 \approx 2.5 z_t - 7.9 \psi_{tx} + 3.4 \psi_{ty} + 5.2 \omega_t + 2.4 \theta_{fl2} - 2.4 \theta_{fr2} + 2.3 \theta_{bl} - 3.9 \theta{bl2} + 2.4 \theta_{br}$}
This expert's scoring function depends on several factors: the z-coordinate of the torso, its orientation, and the state of several joints, which suggests that this expert may be called in a variety of cases.

\begin{equation}
    \begin{cases}
        \tau_{br} \approx 0\\
        \tau_{br2} \approx -1.6 z_t\\
        \tau_{fl} \approx 0\\
        \tau_{fl2} \approx 0\\
        \tau_{fr} \approx -1.8 z_t\\
        \tau_{fr2} \approx 0\\
        \tau_{bl} \approx 4.0 z_t\\
        \tau_{bl2} \approx 0\\
    \end{cases}
\end{equation}
This policy is extremely simpler than those of the other experts, suggesting that the only piece of information used by this expert is the z-coordinate, used in only $3$ cases, while the other joints do not receive any torque.
This suggests that this expert is called in ``transitions'' between one behavior and the other, which allows the robot to exploit its own momentum without any significant change. 

\begin{figure*}[ht!]
\centering
\includegraphics[width=0.95\textwidth]{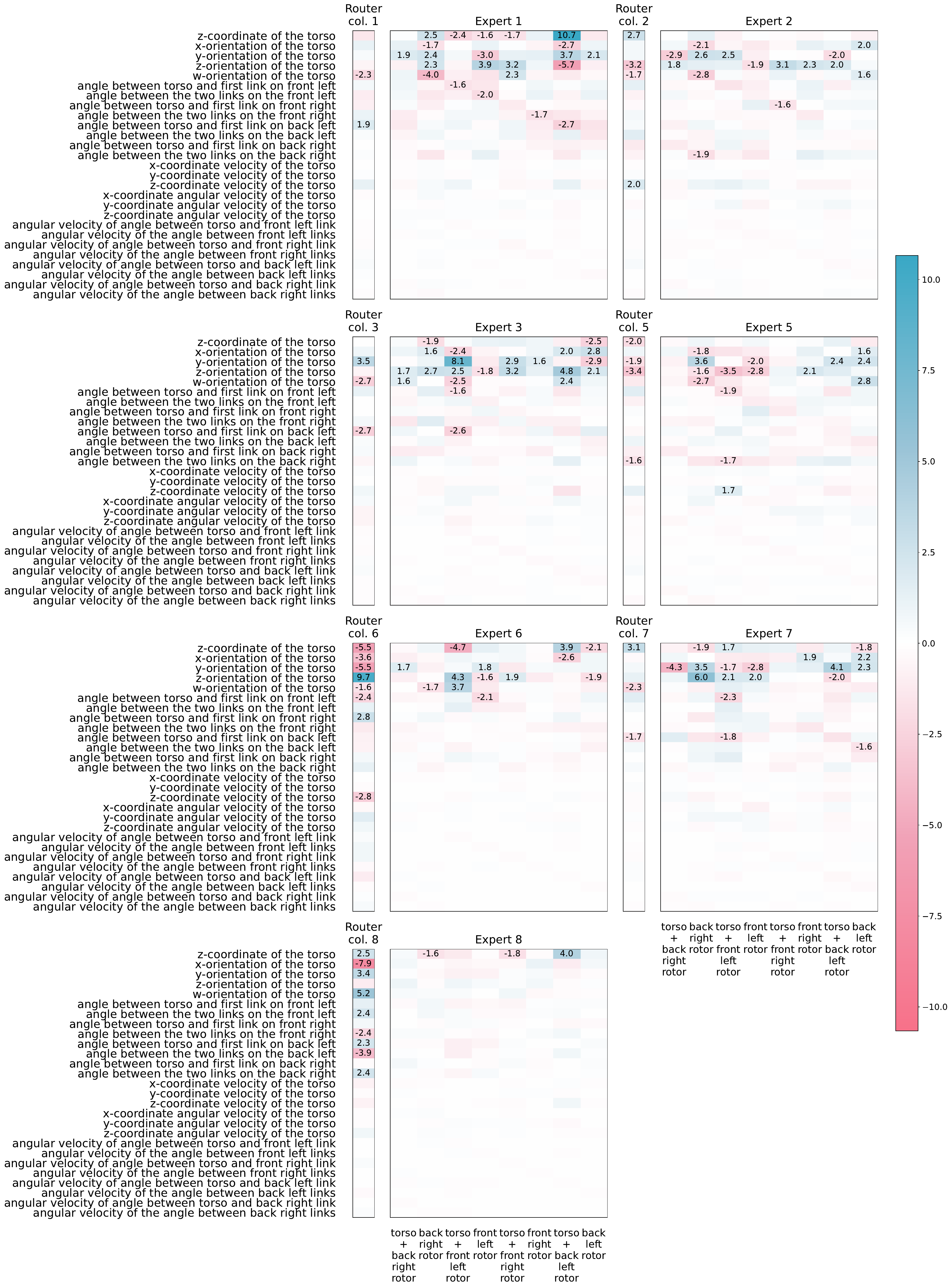}
\caption{\textbf{Ant-v4}. Visual representation of the learned weights for each expert and of the corresponding column of the router’s weight matrix.}
\label{fig:ant_weights}
\end{figure*}

\end{document}